\documentclass[a4paper,fleqn]{cas-sc}
\usepackage[square,numbers]{natbib}
\usepackage{balance}
\usepackage{color,soul}
\usepackage{graphicx}
\usepackage{amstext}
\usepackage{amsmath}
\usepackage{amssymb}
\usepackage{booktabs}
\usepackage{caption}
\usepackage{subcaption}
\usepackage{siunitx}
\usepackage{hyperref}
\usepackage{url}
\usepackage{multirow}
\usepackage{multicol}
\usepackage{algorithm}
\usepackage{algorithmic}
\usepackage{academicons}
\usepackage{xcolor}
\usepackage{xspace}

\def\tsc#1{\csdef{#1}{\textsc{\lowercase{#1}}\xspace}}
\tsc{WGM}
\tsc{QE}
\tsc{EP}
\tsc{PMS}
\tsc{BEC}
\tsc{DE}

\begin{document}

\let\WriteBookmarks\relax
\def\floatpagepagefraction{1}
\def\textpagefraction{.001}

\shorttitle{Involution-Infused DenseNet with Two-Step Compression for Resource-Efficient Plant  Disease Classification.}


\title [mode = title]{Involution-Infused DenseNet with Two-Step Compression for Resource-Efficient Plant  Disease Classification.}                      
\tnotemark[]



\shortauthors{Ahmed et al.}


\author[1, 2]{Tashik Ahmed}[orcid=0009-0001-7973-6795]
\cormark[1] 
\ead{tashik.ahmed@bracu.ac.bd}

\author[1]{Sumaitha Jannat}[orcid=0009-0003-5047-6787]
\ead{sumaitha.jannat@g.bracu.ac.bd}
                   
\author[1, 2]{Md. Farhadul Islam}[orcid=0000-0003-3249-4490]
\ead{farhadulfuad324@gmail.com}

\author[1, 2]{Jannatun Noor}[orcid=0000-0001-9669-151X]
\ead{jannatun@cse.uiu.ac.bd}

\cortext[cor1]{Corresponding Author}

\address[1]{Department of Computer Science and Engineering, School of Data and Sciences, BRAC University, Dhaka, Bangladesh}

\address[2]{Computing for Sustainability and Social Good (C2SG) Research Group, Department of Computer Science and Engineering, United International University, Dhaka, Bangladesh}

\begin{abstract}
Agriculture is vital for global food security, but crops are vulnerable to diseases that impact yield and quality. While Convolutional Neural Networks (CNNs) accurately classify plant diseases using leaf images, their high computational demands hinder their deployment in resource-constrained settings such as smartphones, edge devices, and real-time monitoring systems. This study proposes a two-step model compression approach integrating Weight Pruning and Knowledge Distillation, along with the hybridization of DenseNet with Involutional Layers. Pruning reduces model size and computational load, while distillation improves the smaller student model's performance by transferring knowledge from a larger teacher network. The hybridization enhances the model’s ability to capture spatial features efficiently. These compressed models are suitable for real-time applications, promoting precision agriculture through rapid disease identification and crop management. The results demonstrate ResNet50’s superior performance post-compression, achieving 99.55\% and 98.99\% accuracy on the PlantVillage and PaddyLeaf datasets, respectively. The DenseNet-based model, optimized for efficiency, recorded 99.21\% and 93.96\% accuracy with a minimal parameter count. Furthermore, the hybrid model achieved 98.87\% and 97.10\% accuracy, supporting the practical deployment of energy-efficient devices for timely disease intervention and sustainable farming practices.

\end{abstract}




\maketitle
\section{Introduction}
Agriculture, an indispensable pillar of human civilization, plays a critical role in sustaining life and supporting economies. However, crop diseases remain a persistent threat, degrading crop quality, reducing yields, and undermining food security. Early and accurate disease detection is crucial for preventing major losses, but manual inspection is still the predominant method in many regions. Unfortunately, it is time-consuming, prone to human error, and often results in delayed responses, disrupting crop cycles and threatening the stability of food supply chains.

Automated disease detection systems offer significant benefits over manual inspection by reducing human error and providing faster, more consistent results. These systems, optimized for edge devices like smartphones and drones, give farmers real-time insights to quickly address infestations, minimizing crop damage and promoting sustainable practices. Advanced deep learning models, especially Convolutional Neural Networks (CNNs), excel in plant disease identification with high accuracy. Techniques like knowledge distillation and pruning compress these models, making them suitable for resource-constrained environments without sacrificing performance. Automation and deep learning together enhance agricultural productivity, helping farmers safeguard yields and strengthen global food security.

Research indicates that plant leaf diseases destroy 30-40\% of crop yields annually \cite{r1}, causing over \$40 billion in economic losses. In the last five years, more than 800 million people have faced food insecurity \cite{r2}. Crops like rice \cite{r3}, maize \cite{r4}, and peanuts \cite{r5} are affected by leaf spot, one of over 19,000 fungal diseases threatening food security \cite{r6}. Disease identification typically relies on professionals, increasing costs and errors, complicating detection \cite{r7}. Automated, intelligent detection remains challenging in precision agriculture \cite{r8}, but advancements in deep learning now enhance image classification and object recognition \cite{r9, r10, r11}.

Several studies have explored CNN-based architectures for plant disease detection. One study \cite{r12} compares six models (ResNet152V, DenseNet121, MobileNetV2, Seresnext101, InceptionV3, and ResNext101) and highlights the superior accuracy of an ensemble model (DEX) with transfer learning for rice disease detection, though limited by Colab’s restrictions on optimizers like Adadelta and NAdam. Another study \cite{r13} develops an optimized CNN to classify 38 plant diseases, achieving 95\% accuracy, with weight pruning reducing network size by 66.7\% and increasing unseen data accuracy to 93.9\%. However, pruned models may still be too complex for low-end mobile devices lacking advanced hardware accelerators, leading to slower inference. Additionally, ECA-KDNet \cite{r14} employs knowledge distillation and the ECA attention mechanism to diagnose apple leaf diseases, achieving 98.28\% accuracy with only 3.38M parameters, making it suitable for mobile devices. While knowledge distillation reduces computational demands, further compression could risk slight accuracy trade-offs.

This research is motivated by the need to deploy complex deep-learning models in agricultural environments where real-time processing and efficiency are critical. Although models like ResNet, DenseNet, and VGG-16 achieve high accuracy in plant disease detection and crop monitoring, their significant computational and memory demands make them unsuitable for resource-constrained settings, such as farms relying on mobile devices, drones, or low-power edge devices. These environments often lack access to high-end hardware like GPUs, creating bottlenecks in processing large visual datasets efficiently. To address these challenges, the research focuses on optimizing models like VGG-16, ResNet50, and DenseNet169 by applying Weight Pruning, Knowledge Distillation, and Involution techniques to reduce size and computational complexity while maintaining accuracy. The goal is to create lightweight models that balance performance and efficiency, making them suitable for deployment on edge devices. This will enable real-time disease detection and monitoring, enhancing agricultural productivity and decision-making in dynamic environments. In recent years, vision tasks in healthcare and agriculture have focused on such resource-constrained implementations \cite{r15, r16, r17, r18}. Our proposed method combines the advantages of network pruning, knowledge distillation with an efficient compact model.

Researchers have developed various CNN-based models for plant leaf disease detection, achieving high accuracy. However, models like VGG16, ResNet50, and DenseNet169 are computationally intensive due to their numerous layers, making them challenging to deploy on edge devices. This research addresses these limitations by integrating weight pruning, which reduces computational complexity by setting certain neural network weights to zero. It also implements knowledge distillation, enabling smaller models to learn from larger, complex ones. Additionally, the involution process is applied with DenseNet, utilizing fewer parameters and a dynamic kernel to extract locational information often overlooked by standard convolution, thereby enhancing accuracy and precision. The goal is to optimize these large CNN architectures to ensure compatibility with edge devices for real-time applications. By deploying these models on low-power devices such as smartphones or drones, the research aims to make plant disease detection easier and more accessible for farmers, promoting timely interventions and improved crop management.

Drawing from our research, we have the following contributions to present by this study:
\begin{itemize}
    \item We introduce a dual-step model compression approach combining knowledge distillation and weight pruning, resulting in a significant compression in size for various CNN models.
    \item We implement an Involution-infused DenseNet student model that demonstrates enhanced performance and robustness under noisy conditions with fewer parameters.
    \item This research offers an adaptation of CNN Models for real-time deployment by tailoring compressed versions of VGG16, ResNet50, DenseNet169, MobileNetV2, and EfficientNet-B0.
    \item Our proposed methodology demonstrates a significant improvement in resource efficiency while maintaining inference time, advancing real-time disease monitoring in resource-constrained agricultural environments.
\end{itemize}
\section{Related Works}
\label{Related Works}
\subsection{Plant Leaf Disease and Image Classification using CNN}
An ensemble model, DEX, combining XceptionNet, EfficientNetB7, and DenseNet121, achieved 98\% accuracy in detecting common rice diseases in Bangladesh. Among other CNN architectures tested, Inceptionv3 and DenseNet101 showed the highest accuracy at 97\%, while SeresNext101 initially performed the lowest at 79\%, but improved to 96\% with transfer learning \cite{r12}. The dataset, expanded from 900 to 42,876 images, was used effectively for training and testing. Another study introduces a hybrid CNN-transformer architecture, FOTCA, utilizing Adaptive Fourier Neural Operators (AFNO), which achieved the highest accuracy (99.8\%), surpassing models like DenseNet169 and ResNet50. FOTCA uses Focal Loss to better handle challenging samples, combining local and global features for improved performance. This model was trained and validated on the PlantVillage dataset, consisting of 54,303 images across 38 species \cite{r19}. 

Additionally, research developed an automated system for classifying medicinal plants using a CNN model with three convolutional layers and data augmentation techniques. Trained on 34,123 images and tested on 3,570, the model achieved 71.3\% accuracy, using a dataset of 10 medicinal plants from various regions \cite{r20}. Another study tackled the challenge of plant species identification by proposing a multi-deep, multi-path convolutional network, which captures diverse plant image areas for improved feature extraction. Experimental results confirmed its effectiveness in plant classification \cite{r21}. Further research applied TensorFlow and Keras to train models on corn images from the PlantVillage dataset using DenseNet201 and custom models with Stochastic Depth and Dropout, outperforming architectures like VGG16, InceptionV3, and ResNet50. With a learning rate of 0.2\%, 35\% dropout, and the Adam optimizer, the model achieved 98.36\% testing accuracy after 30 epochs. Transfer learning and a two-stage training process enhanced the model’s performance in recognizing corn diseases \cite{r22}.
\subsection{Weight Pruning and Knowledge Distillation}
A study proposes an "Entropy-based pruning" method for CNNs, dynamically adjusting pruning thresholds during training based on weight contributions, improving compression \cite{r23}. Applied to LeNet-5 on the MNIST dataset, it achieves a 28.25\% higher compression rate without loss of accuracy. Then, another study proposes combining Knowledge Distillation with a pruning filter to compress CNN models, which is applied to MobileNets. After reducing the model size by 26.08\%, accuracy increases from 63.65\% without KD to 65.37\% with KD, though the model without KD had a 0.1s faster inference time \cite{r24}. A study develops a CNN architecture with minimal parameters, using six two-dimensional convolutional layers and dropout to prevent over-fitting. The model, optimized with Adam and a 0.001 learning rate, achieves 98.18\% accuracy with enhanced anomaly detection and computational efficiency over 50 epochs \cite{r25}. Another paper proposes a lightweight convolutional neural network model named PLeaD-Net for plant leaf disease classification. It achieves a testing accuracy of 98.18\% on the PlantVillage dataset, with significantly fewer parameters compared to other deep learning models. This makes it highly efficient regarding computational resources \cite{r26}.
\section{Precursors}
\label{Precursors}
\subsection{CNN in classifying Plant Leaf Diseases}
Convolutional Neural Networks (CNNs) are specialized deep learning models designed to process structured raster data, such as images. They consist of layers with learnable filters or kernels that convolve input data to detect local patterns and features. In plant leaf disease classification, CNNs can automatically learn hierarchical representations, starting with simple features like edges, textures, and color variations, and progressing to complex structures such as disease-specific lesions, spots, or discolorations. Pooling layers reduce the spatial dimensions, enabling the network to focus on the most critical features, such as areas affected by diseases, while ignoring irrelevant background information. This ability to capture spatial hierarchies and maintain transformation invariance makes CNNs particularly effective in plant disease detection. By learning from annotated leaf images, CNNs can classify various diseases with high accuracy, facilitating early identification and enabling farmers to take timely interventions. CNN architectures have thus become essential in automating disease recognition and advancing agricultural diagnostics. We use 5 CNN architectures, which are VGG-16, ResNet50, DenseNet169, MobileNetV2, and EfficientNet-B0. VGG-16 provides a clear grasp of the characteristics of an image, including its shape, textures, and patterns. ResNet50 used residual blocks to resolve the Vanishing Gradient issue. Using the dense connection, DenseNet169 utilizes inputs from all the previous layers, which makes it efficient for feature reuse. A lighter CNN model, MobileNetV2, is tailored for mobile and resource-constrained environments. Last but not least, EfficientNet-B0 offers a scalable and efficient solution by uniformly balancing the depth, width, and resolution of the network using a novel compound scaling method. Though CNN looks like a solution to the classification problems, the computational cost of this architecture restrains its deployment in resource-constrained devices. The memory demands of these architectures hinder their practical use in autonomous farming systems.
\subsection{Optimization Techniques for CNN}
\subsubsection{Weight Pruning}
Weight pruning in Convolutional Neural Networks (CNNs) reduces model parameters by removing less important weights, resulting in smaller, faster, and more efficient networks with minimal performance impact. This technique creates a sparse network, typically through unstructured pruning, which sets individual weights to zero rather than removing entire nodes. The challenge lies in identifying which weights to prune, often determined by their significance using heuristics based on weight magnitude. Pruning is generally performed after training; however, it may lead to a decline in model performance, which can be mitigated through fine-tuning, retraining the model post-pruning to regain accuracy. The necessity for fine-tuning and the extent of pruning can vary based on the specific use case and applied techniques. This corresponds to the following equation:
\begin{equation}
\min_{\mathbf{W}'} \| C(D \mid \mathbf{W}') - C(D \mid \mathbf{W}) \| \quad \text{s.t.} \quad \| \mathbf{W}' \|_0 \leq B
\end{equation}
Where the $||W^{'}||_0$ bounds the number of non-zero parameters to $B$ in $W^{'}$. $W$ represents the parameters in the network. By $C$, it means Cost Function.
\subsubsection{Knowledge Distillation}
Knowledge distillation is a machine learning technique that transfers knowledge from a larger, complex model (teacher) to a smaller, efficient model (student), as introduced by Geoffrey Hinton \cite{r27}. In Convolutional Neural Networks (CNNs), the teacher model is a high-accuracy, extensively trained CNN, while the student model aims to replicate its performance with fewer parameters and computational resources. Instead of using hard labels for training, the student model learns from soft targets—probability distributions predicted by the teacher model. These soft targets provide more informative supervision signals by revealing class relationships. The training of the student model typically involves a distillation loss, which measures the discrepancy between its predictions and the soft targets generated by the teacher model. A common form of distillation loss is the Kullback-Leibler (KL) divergence between the softmax outputs of the teacher and student models:
\begin{equation}
    \mathcal{L}_{\text{distill}} = \text{KL}(P_{\text{teacher}} \| P_{\text{student}})
\end{equation}
where $P_{teacher}$ and $P_{student}$ are the softmax outputs of the teacher and student models, respectively.
\subsubsection{Involution Process}
\begin{figure}[h]
    \centering
    \includegraphics[width=\linewidth]{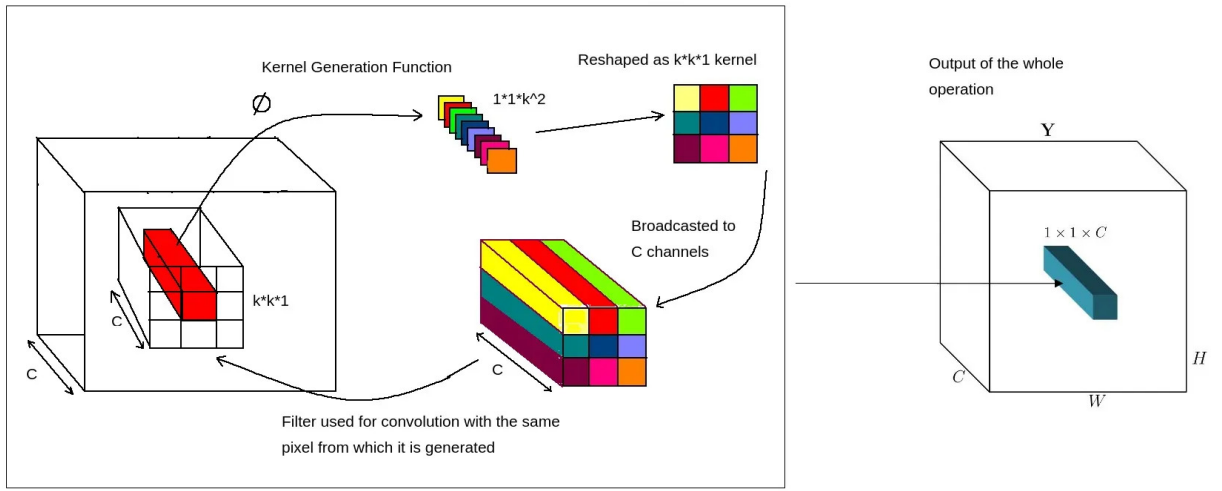}
    \caption{Involution process \cite{r28}}
    \label{invo}
\end{figure}
The involution process \cite{r28}, a variant of convolution, is beneficial for classification due to its fewer weight parameters and dynamic kernel generation that captures locational information. It effectively addresses inter-positional dependencies using a single set of meta-weights, allowing for larger models with improved performance and reduced memory usage. Our study shows that incorporating even one involution layer into a CNN significantly enhances classification accuracy and recall by capturing long-range dependencies and complex spatial interactions. However, excessive use may lead to overfitting. Therefore, a balanced approach that combines conventional convolution with involution layers is recommended for optimal performance.

\section{Dataset}
\label{dataset}
We use two datasets.  The first dataset is the PlantVillage Dataset \cite{r29}. It comprises 54,306 plant leaf images. It contains a total of 38 classes and samples of 14 different species. Here, 26 classes contain samples of diseased plants. The 12 remaining classes are of healthy plants. This dataset serves as an efficient resource for detecting different plant diseases. Our source of data was Kaggle. To improve crop management practices, this dataset serves an excellent purpose for our research, which is disease detection. In this dataset, we have diseases of Apple, Blueberry, Cherry, Corn, Grape, Orange, Peach, Raspberry, Soybean, Squash, Strawberry, Pepper bell, Potato, and Tomato. The dataset has been loaded with the "ImageFolder" class from the PyTorch library. 80\% of the data was allocated for training and the remaining 20\% for testing.
\begin{figure}[htbp]
    \centering
    \begin{subfigure}[t]{0.25\textwidth}
        \centering
        \includegraphics[width=\linewidth]{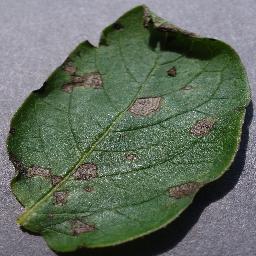}
        \caption{PlantVillage}
        \label{fig:fig1}
    \end{subfigure}
    \hspace{0.02\textwidth}
    \begin{subfigure}[t]{0.25\textwidth}
        \centering
        \includegraphics[width=\linewidth]{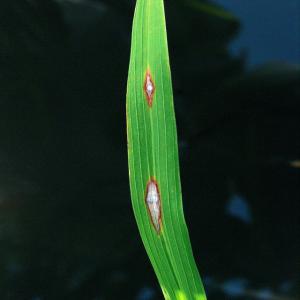}
        \caption{Paddy Leaf}
        \label{fig:fig2}
    \end{subfigure}
    \caption{Sample images from the PlantVillage dataset \cite{r29} and the Paddy Leaf dataset \cite{r30}}
    \label{fig:twofigs}
\end{figure}

The second dataset used is the Paddy Leaf dataset \cite{r30}, consisting of approximately 6000 samples distributed across four classes: Bacterial Blight, Blast, Brown Spot, and Tungro. The dataset is split into training (70\%), validation (15\%), and test (15\%) sets. We analyze the outcomes of both Knowledge Distillation and Weight Pruning, focusing on model size, accuracy, inference time, and other evaluation metrics such as precision, recall, and F1 score. Hyperparameters for the experiments include a batch size of 32, a learning rate of 0.001, and the Adam optimizer.
\section{Methodology}
\label{meth}
In this part, we illustrate the workflow, details of the two-step compression pipeline, and the proposed model architecture.
\subsection{Workflow of the study}
\begin{figure}[ht]
    \centering
    \includegraphics[width=0.8\linewidth]{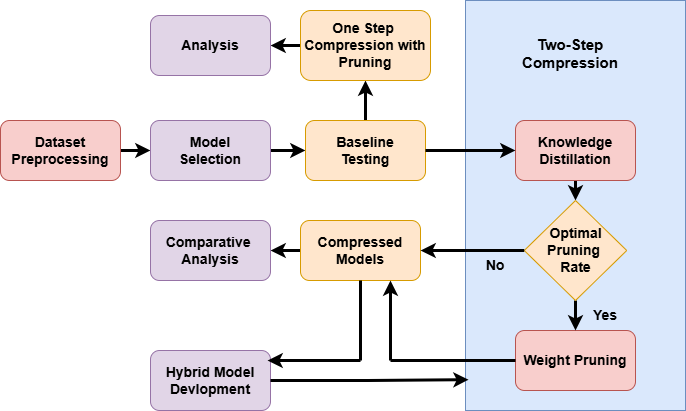 }
    \caption{Workflow of our study}
    \label{fig: wf}
\end{figure}
Initially, after identifying the problem in this research, we study the relevant works related to the problem in order to identify the gains and gaps. Secondly, we collect suitable datasets (PlantVillage and Paddy Leaf) that align with our problem. After dataset collection, we apply the necessary preprocessing steps and make the datasets ready for deployment. In the next step, we select several suitable CNN models for the research. The selected baseline models are VGG16, ResNet50, DenseNet169, EfficientNet-B0, and MobileNet V2. Then we implement baseline testing over the selected datasets. After evaluating the baseline results, we forward the models through one-step compression and two-step compression, respectively, and do a comparative analysis of the compressed models. Lastly, we develop the hybrid model by infusing an involutional layer and evaluating its performance. After each operational step, we checked for the overall optimized performance of the model and applied necessary modifications based on that information. Figure \ref{fig: wf} demonstrates the workflow.
\subsection{The Two-Step compression Pipeline}
\begin{figure}[ht]
    \centering
    \includegraphics[width=0.9\linewidth]{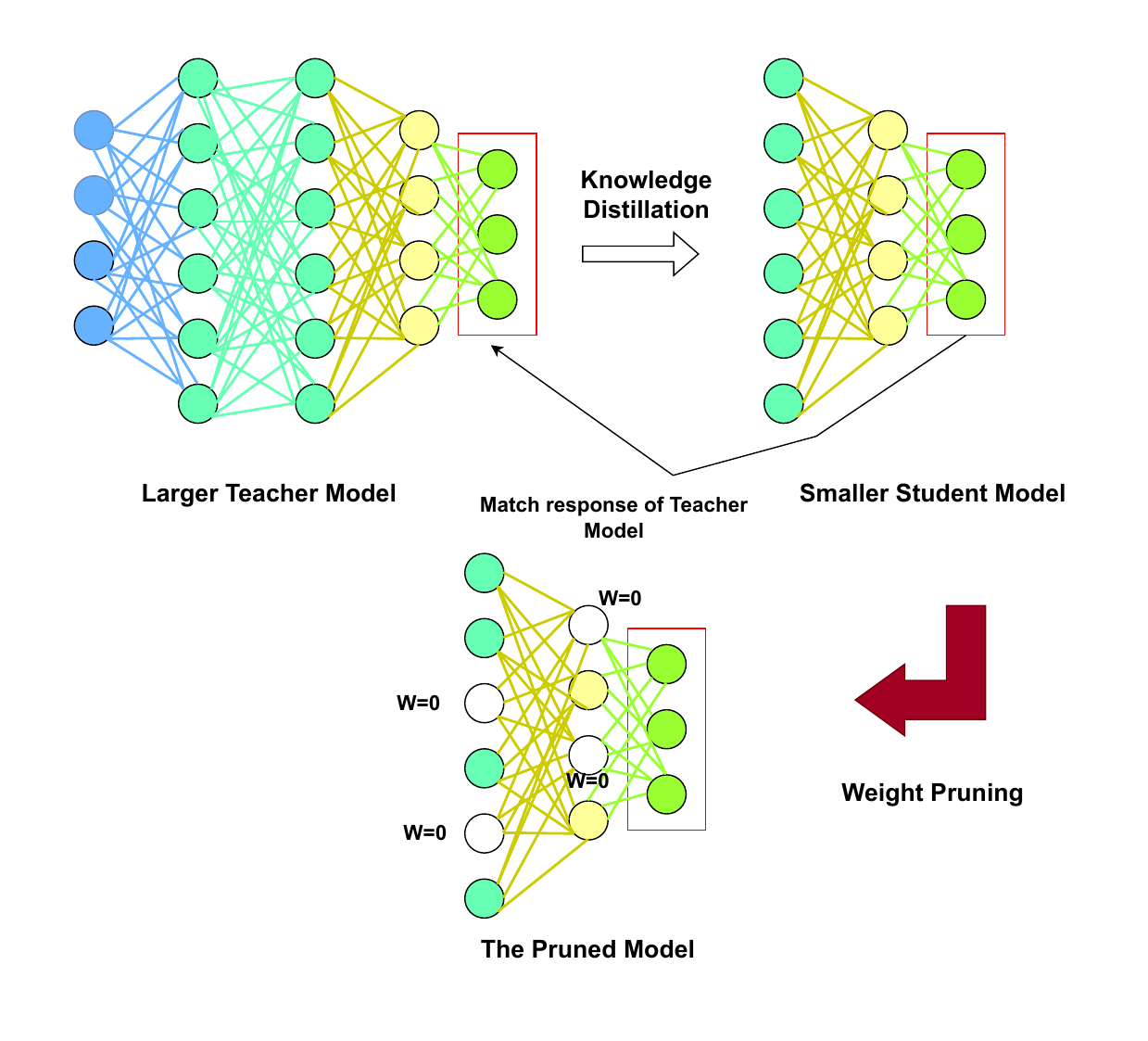}
    \caption{The architecture of the two-step compression pipeline}
    \label{fig: 2s}
\end{figure}

This study builds upon our previously proposed two-step compression framework for Convolutional Neural Networks (CNNs) \cite{r31}, which integrates knowledge distillation with
post-training weight pruning to develop lightweight and efficient models for rice leaf disease classification. We briefly revisit the methodology here to establish continuity before presenting our extended contributions. As illustrated in Figure \ref{fig: 2s}, the process begins with a teacher model, pre-trained on a larger dataset, transferring its learned knowledge to a more efficient student model via knowledge distillation. In the next step, selective threshold-based weight pruning is applied to specific layers of the student model, enhancing its efficiency without compromising accuracy. The final pruned student model is then evaluated and fine-tuned if necessary to ensure an optimal balance between performance and computational efficiency.
The pipeline begins with a pre-trained teacher model transferring knowledge \cite{r32} to a compact
student model via a custom distillation loss. This loss combines a soft loss, computed using Kullback-Leibler (KL) divergence \cite{r33} between softened output distributions of the teacher and student, and a hard loss, measured by cross-entropy with the ground truth labels. Temperature
scaling is applied to logits to refine probability distributions, and the two loss components are
balanced via a tunable parameter, $alpha$, as shown below:
\begin{itemize}

\item Temperature scaling:
\begin{equation}
    Z_T(T)=\frac{Z_T}{T}
\end{equation}
\begin{equation}
     Z_S(T)=\frac{Z_S}{T}
\end{equation}
\item Soft Loss (KL Divergence):
\begin{equation}
    L_{Soft}=T^{2}\cdot KL(Softmax(\frac{Z_T}{T})|| log(Softmax(\frac{Z_S}{T})))
\end{equation}
\item Hard Loss Cross Entropy:
\begin{equation}
    L_{Hard} = CrossEntropy (Z_S,y)
\end{equation}
\item Total Loss:
\begin{equation}
    L_{Total} = (1 - \alpha)\cdot L_{Hard} + \alpha \cdot L_{Soft}
\end{equation}
\end{itemize}

For student model construction, we apply architectural reductions while preserving core principles of the 
teacher networks. For example, the compressed VGG16 reduces filter sizes to (16, 32, 64) and downsizes 
fully connected layers to 512 neurons. The ResNet50 student halves the filters and reorganizes blocks to 
(2, 2, 4, 2), maintaining residual connections. The DenseNet169 student scales down to three 
DenseBlocks with [3, 4, 5] layers, a growth rate of 32, and transition layers that downsample and restore 
channel dimensions to 16. Additionally, we introduce two new students: an EfficientNet-B0 variant with 
reduced width (16 to 256 channels) across four convolutional blocks and a MobileNetV2 student with a 
full-width multiplier (1.0), implementing four custom bottleneck blocks to increase depth. These models 
aim to retain performance while enabling effective compression. 
Following distillation, we employ a post-training pruning approach \cite{r34}, where weights are pruned after the model is fully trained. Weights below a layer-wise threshold $\tau$ (based on percentile ppp) are zeroed out, excluding the last layer, bias, and batch normalization parameters. The mathematical formulation is as follows:
\par Let $W$ $\in$ $\mathbb{R}^{n*m}$ represent the weight matrix of a particular layer. $p$ is the pruning percentage, $\tau$ is the pruning threshold, determined by the $p$-th percentile of the absolute values of the weight in $W$. $M$ $\in$ $\{0,1\}^{n*m}$ is the binary mask for the layer. $W'$ is the pruned weight matrix after applying the mask. $|W_{i,j}|$ refers to the absolute value of the weight at position $(i,j)$ in the matrix.
\begin{algorithm}[!ht]
\caption{Weight Pruning with Masking}
\label{alg:weight_pruning}
\begin{algorithmic}[1]
\STATE \textbf{Input:}
\STATE \quad $M$: pre-trained model
\STATE \quad $p$: Pruning percentage
\STATE \quad $D$: Dataloader (for evaluation)

\STATE Initialize $M' \gets \text{deepcopy}(M)$
\STATE Initialize $masks \gets []$

\FOR{each parameter $\theta$ in $M'$ (excluding biases and final layers)}
    \STATE $W \gets \theta$ \COMMENT{weights as a numpy array}
    \STATE $\tau \gets \text{Percentile}(|W|, p)$ \COMMENT{Pruning threshold}
    \STATE $m \gets W \geq \tau$ \COMMENT{Binary mask (1 if $|W| \geq \tau$, else 0)}
    \STATE $W \gets W \odot m$ \COMMENT{Apply mask (prune weights below threshold)}
    \STATE Update $\theta$ with pruned $W$
    \STATE $masks.\text{append}(m)$
\ENDFOR

\STATE $acc \gets \text{Evaluate}(M', D)$ \COMMENT{Accuracy after pruning}
\STATE $t_i \gets \text{Measure\_inference\_time}(M', D)$ \COMMENT{Inference time after pruning}

\STATE \textbf{Return:}
\STATE \quad $(M', masks, acc, t_i)$

\STATE \textbf{Output:}
\STATE \quad $M'$: Pruned model
\STATE \quad $masks$: Masks for pruned layers
\STATE \quad $acc$: Accuracy of pruned model
\STATE \quad $t_i$: Inference time of pruned model
\end{algorithmic}
\end{algorithm}
To calculate the threshold,
\begin{equation}
    \tau= Percentile(|W|,p)
\end{equation}
This defines the pruning threshold $\tau$, below which the weights will be pruned.
For the Mask,
\[
M_{i,j} = 
\begin{cases} 
1 & \text{if } |W_{i,j}| \geq \tau \\
0& \text{if } |{W}_{i,j}| < \tau
\end{cases}
\]
This Mask $M$ will retain the weights greater than or equal to the threshold $\tau$, while zeroing out the others.
Now, we compute the pruned weight matrix:
\[
W'= W \odot M
\]
The pruned weight matrix $W'$ is obtained by the element-wise multiplication of the mask $M$ with the weight matrix $W$.
The final formula is as follows:
\[
W'_{i,j} = 
\begin{cases} 
W_{i,j} & \text{if } |W_{i,j}| \geq \tau \\
0& \text{if } |{W}_{i,j}| < \tau
\end{cases}
\]
This pruning approach is iteratively evaluated across various pruning percentages (0\%–99\%) to determine the optimal trade-off between model efficiency and classification performance.
\subsection{The Proposed Model}
\begin{figure}[]
    \centering
    \includegraphics[width=\linewidth]{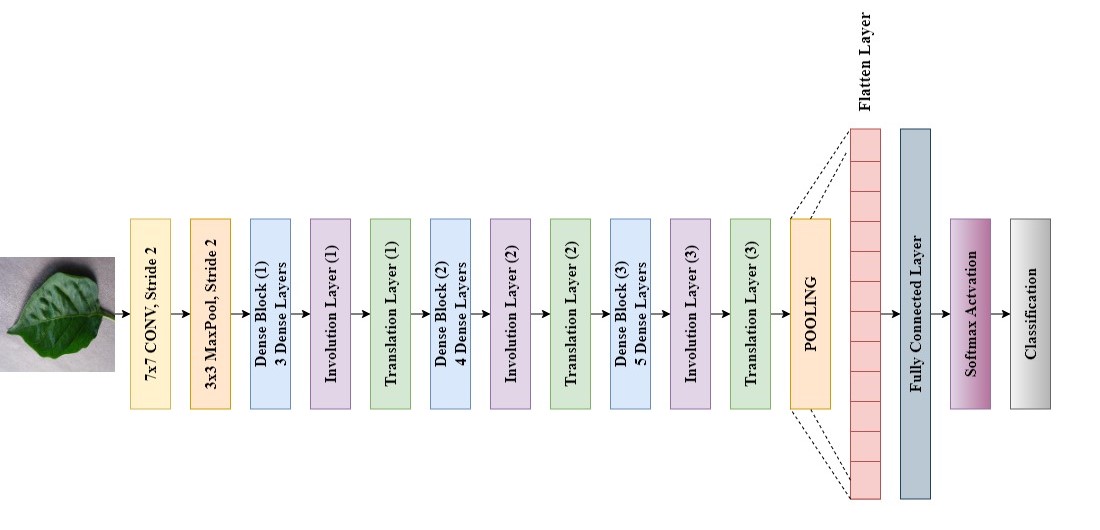}
    \caption{The proposed model architecture}
    \label{fig: prop}
\end{figure}
The proposed model, illustrated in Figure \ref{fig: prop}, is a compressed DenseNet-like architecture that combines DenseBlocks and Involution Layers for enhanced feature extraction with minimal parameters. It starts with a 7x7 convolutional layer for RGB input images, followed by batch normalization and ReLU activation. A max-pooling layer reduces the spatial dimensions of the feature maps. The model contains three sequential DenseBlocks, with the first block comprising 3 layers, the second 4 layers, and the third 5 layers, each utilizing a bottleneck structure with a 1x1 convolution followed by a 3x3 convolution to ensure feature reuse. After each DenseBlock, an Involution layer dynamically generates convolutional kernels based on the input, using a reduction ratio of 2 for efficient filtering. Transition layers following each Involution layer reduce channel and spatial dimensions. The model concludes with a global average pooling layer, flattening the output for a fully connected layer that produces predictions, totaling 0.32 million parameters and requiring 1.22 MB of memory.
\section{Results} \label{results}
Our experiments are performed using Google Colab Pro. We use an Intel Core i5 CPU, which provides reliable processing power for various applications. The system is equipped with an NVIDIA T4 GPU (for the Paddy Leaf) and an A100 GPU (for the PlantVillage Dataset), suitable for tasks requiring enhanced graphics and computational capabilities. Storage is well-managed with a combination of a 1 TB HDD and a 256 GB SSD, ensuring ample space and faster data access. Additionally, the system includes 8 GB of RAM, which supports smooth multitasking and efficient performance. About hyperparameters, the learning rate is 0.001, epochs are around 20-25. In the case of Knowledge Distillation, alpha is 0.5, and the temperature is 3.00-5.00.
\subsection{Performance Analysis of Baseline Models}
\begin{table}[]
\caption{Results of baseline models}
\resizebox{\linewidth}{!}{
\begin{tabular}{|c|c|ccccc|}
\hline
\multirow{2}{*}{\textbf{Dataset}}                                                     & \multirow{2}{*}{\textbf{Parameters}}                                      & \multicolumn{5}{c|}{\textbf{Models}}                                                                                                                      \\ \cline{3-7} 
                                                                                      &                                                                           & \multicolumn{1}{c|}{VGG-16} & \multicolumn{1}{c|}{ResNet50} & \multicolumn{1}{c|}{DenseNet169}    & \multicolumn{1}{c|}{MobileNet\_V2} & EfficientNet\_B0 \\ \hline
\multirow{2}{*}{\begin{tabular}[c]{@{}c@{}}PlantVillage\\ \&\\ RiceLeaf\end{tabular}} & \begin{tabular}[c]{@{}c@{}}MACS\\ (\num{e9})\end{tabular}   & \multicolumn{1}{c|}{15.47}  & \multicolumn{1}{c|}{4.10}     & \multicolumn{1}{c|}{3.38}           & \multicolumn{1}{c|}{\textbf{0.31}}          & 0.39             \\ \cline{2-7} 
                                                                                      & \begin{tabular}[c]{@{}c@{}}FLOPS\\ (\num{e9})\end{tabular}  & \multicolumn{1}{c|}{30.94}  & \multicolumn{1}{c|}{8.20}     & \multicolumn{1}{c|}{6.76}           & \multicolumn{1}{c|}{\textbf{0.61}}          & 0.79             \\ \hline
\multirow{5}{*}{PlantVillage}                                                         & \begin{tabular}[c]{@{}c@{}}Accuracy\\ (\%)\end{tabular}                   & \multicolumn{1}{c|}{90.79}  & \multicolumn{1}{c|}{99.13}    & \multicolumn{1}{c|}{99.40}          & \multicolumn{1}{c|}{99.67}         & \textbf{99.78}   \\ \cline{2-7} 
                                                                                      & \begin{tabular}[c]{@{}c@{}}Average \\ Inference Time\\ (sec)\end{tabular} & \multicolumn{1}{c|}{0.18}   & \multicolumn{1}{c|}{0.13}     & \multicolumn{1}{c|}{0.16}           & \multicolumn{1}{c|}{0.15}          & \textbf{0.13}    \\ \cline{2-7} 
                                                                                      & Precision                                                                 & \multicolumn{1}{c|}{0.99}   & \multicolumn{1}{c|}{0.91}     & \multicolumn{1}{c|}{1.00}           & \multicolumn{1}{c|}{1.00}          & \textbf{1.00}    \\ \cline{2-7} 
                                                                                      & Recall                                                                    & \multicolumn{1}{c|}{0.90}   & \multicolumn{1}{c|}{0.99}     & \multicolumn{1}{c|}{1.00}           & \multicolumn{1}{c|}{1.00}          & \textbf{1.00}    \\ \cline{2-7} 
                                                                                      & F1-score                                                                  & \multicolumn{1}{c|}{0.90}   & \multicolumn{1}{c|}{0.99}     & \multicolumn{1}{c|}{1.00}           & \multicolumn{1}{c|}{1.00}          & \textbf{1.00}    \\ \hline
\multirow{5}{*}{RiceLeaf}                                                             & \begin{tabular}[c]{@{}c@{}}Accuracy\\ (\%)\end{tabular}                   & \multicolumn{1}{c|}{88.14}  & \multicolumn{1}{c|}{99.12}    & \multicolumn{1}{c|}{\textbf{99.55}} & \multicolumn{1}{c|}{98.35}         & 99.45            \\ \cline{2-7} 
                                                                                      & \begin{tabular}[c]{@{}c@{}}Average \\ Inference Time\\ (sec)\end{tabular} & \multicolumn{1}{c|}{0.091}  & \multicolumn{1}{c|}{0.092}    & \multicolumn{1}{c|}{\textbf{0.11}}  & \multicolumn{1}{c|}{0.091}         & 0.095            \\ \cline{2-7} 
                                                                                      & Precision                                                                 & \multicolumn{1}{c|}{0.85}   & \multicolumn{1}{c|}{0.99}     & \multicolumn{1}{c|}{\textbf{0.99}}  & \multicolumn{1}{c|}{0.97}          & 0.99             \\ \cline{2-7} 
                                                                                      & Recall                                                                    & \multicolumn{1}{c|}{0.85}   & \multicolumn{1}{c|}{0.99}     & \multicolumn{1}{c|}{\textbf{1.00}}  & \multicolumn{1}{c|}{0.98}          & 0.99             \\ \cline{2-7} 
                                                                                      & F1-score                                                                  & \multicolumn{1}{c|}{0.84}   & \multicolumn{1}{c|}{0.99}     & \multicolumn{1}{c|}{\textbf{0.99}}  & \multicolumn{1}{c|}{0.98}          & 0.99             \\ \hline
\end{tabular}}
\label{tab1}
\end{table}
Table \ref{tab1} compares baseline model performance on the Paddy-Leaf and Plant Village datasets, highlighting differences in accuracy, inference time, and computational cost (FLOPs). VGG16 performs the worst, with accuracies of 90.79\% and 88.14\%, the slowest inference times (0.18s and 0.091s), and the highest FLOPs at 30.94 B. ResNet50 and DenseNet169 achieve strong accuracy on the Paddy-Leaf dataset (99.13\% and 99.4\%) with moderate FLOPs of 8.20B and 6.76B, respectively. MobileNet V2 and EfficientNet B0 deliver the best results, with 99.67\% and 99.78\% accuracy, perfect classification metrics, and very low FLOPs (0.61B and 0.79B). On the Plant Village dataset, DenseNet169 and EfficientNet B0 again lead (99.55\% and 99.45\%), while MobileNet V2 offers a strong balance of 98.35\% accuracy and low computational cost. Overall, EfficientNet B0 and MobileNet V2 combine high accuracy with minimal FLOPs among the base models.
\begin{figure}[ht]
    \centering
    \begin{subfigure}{0.48\linewidth}
        \centering
        \includegraphics[width=\linewidth]{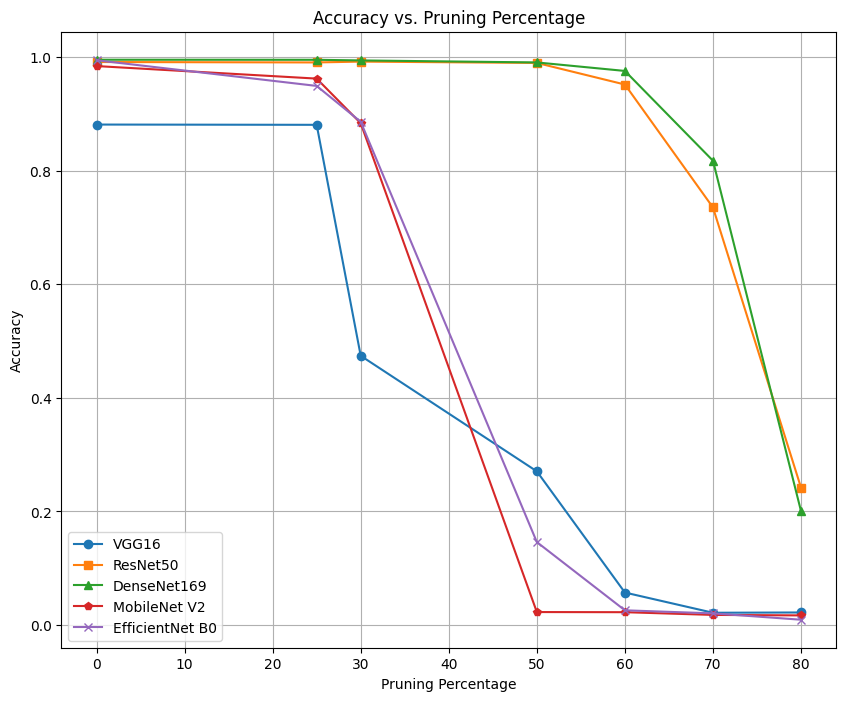}
        \caption{Accuracy vs pruning percentage}
    \end{subfigure}
    \hspace{0.5em}
    \begin{subfigure}{0.48\linewidth}
        \centering
        \includegraphics[width=\linewidth]{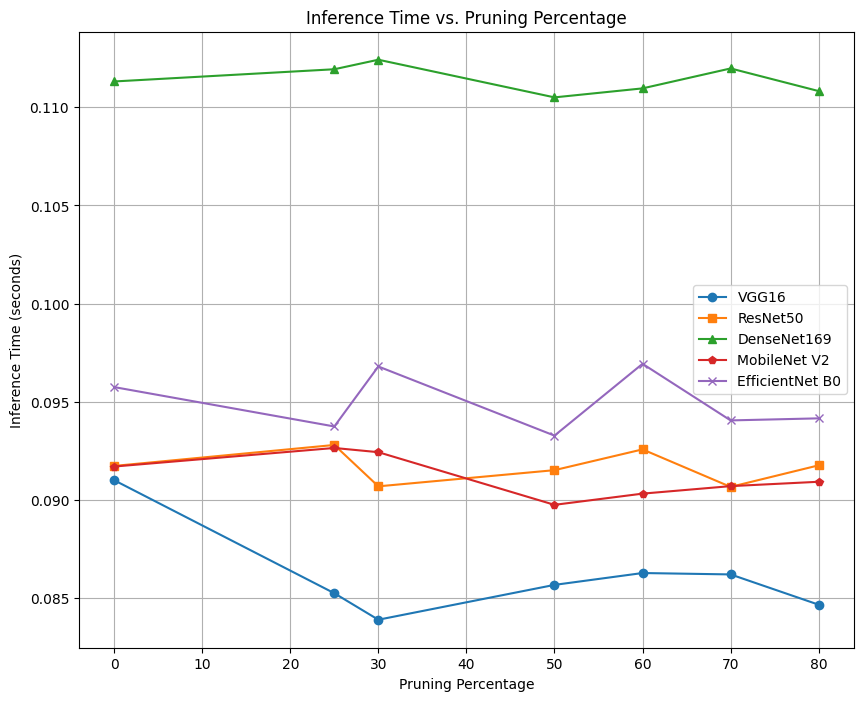}
        \caption{Inference time vs pruning percentage}
    \end{subfigure}
    \caption{Accuracy and inference time vs pruning percentage curves for the PlantVillage dataset}
    \label{c1}
\end{figure}
\begin{table}[ht]
\centering
\caption{Results of baseline models after post-training pruning}
\resizebox{\linewidth}{!}{
\begin{tabular}{cccccccc}
\hline
\textbf{Dataset}              & \textbf{Models} & \textbf{\begin{tabular}[c]{@{}c@{}}Optimal Pruning\\ Rate (\%)\end{tabular}} & \textbf{\begin{tabular}[c]{@{}c@{}}No. of Weighted \\ Parameters\end{tabular}} & \textbf{\begin{tabular}[c]{@{}c@{}}Accuracy\\ (\%)\end{tabular}} & \textbf{\begin{tabular}[c]{@{}c@{}}Avg Inference\\ Time (sec)\end{tabular}} & \textbf{Recall} & \textbf{F1-Score} \\ \hline
\multirow{5}{*}{Paddy Leaf}   & VGG16           & 25                                                                           & 101M                                                                  & 89.33                                                            & 0.16                                                                        & 0.90            & 0.90              \\
                              & ResNet50        & 50                                                                           & 11.78M                                                                & 96.85                                                            & 0.13                                                                        & 0.96            & 0.96              \\
                              & DenseNet169     & 40                                                                           & 7.6M                                                                  & 99.32                                                            & 0.13                                                                        & 0.99            & 0.99              \\
                              & MobileNetV2     & 20                                                                           & 1.79M                                                                 & 98.65                                                            & 0.11                                                                        & 0.99            & 0.99              \\
                              & EfficientNet-B0 & \textbf{20}                                                                           & \textbf{3.22M}                                                                 & \textbf{99.77}                                                            & \textbf{0.12}                                                                        & \textbf{1.00}            & \textbf{1.00}              \\ \hline
\multirow{5}{*}{PlantVillage} & VGG16           & 25                                                                           & 101M                                                                  & 88.62                                                            & 0.084                                                                       & 0.84            & 0.84              \\
                              & ResNet50        & 50                                                                           & 11.85M                                                                & 98.95                                                            & 0.091                                                                       & 0.98            & 0.98              \\
                              & DenseNet169     & 50                                                                           & 6.2M                                                                  & 99.11                                                            & 0.11                                                                        & 0.99            & 0.99              \\
                              & MobileNetV2     & 10                                                                           & 2.05M                                                                 & 98.3                                                             & 0.091                                                                       & 0.99            & 0.99              \\
                              & EfficientNet-B0 & \textbf{15}                                                                           & \textbf{3.46M}                                                                 & \textbf{99.98}                                                            & \textbf{0.094}                                                                       & \textbf{0.99}            & \textbf{0.99}             \\ \hline
\end{tabular}}
\label{tab: t2}
\end{table}

Then, we evaluate the impact of weight pruning on baseline models by analyzing pruning rates from 0\% to 80\% regarding accuracy and inference time across the Paddy-Leaf and Plant Village datasets. Figure \ref{c1}  illustrates the accuracy and inference time changes with varying pruning rates, helping identify optimal rates that balance parameter reduction with performance. Table \ref{tab: t2} shows the details of the performance post-pruning. For the Paddy-Leaf dataset, VGG16 allowed a 25\% pruning, reducing parameters to 101 million while slightly lowering accuracy and improving inference time. ResNet50 achieved 50\% pruning, reducing parameters to 11.78 million with 96.85\% accuracy. DenseNet169, pruned by 40\%, maintained 99.32\% accuracy with 7.6 million parameters and a low inference time of 0.13 seconds. MobileNet V2 and EfficientNet B0, pruned by 20\%, achieved accuracies of 98.65\% and 99.77\%, respectively. In the Plant Village dataset, DenseNet169 again excelled with 50\% pruning, achieving 99.11\% accuracy and 6.2 million parameters.
\subsection{Performance Analysis of Two-Step Compression}
\begin{table}[ht]
\centering
\caption{Results after the first step of the two-step compression}

\begin{tabular}{ccccccc}
\hline
\textbf{Dataset} & \textbf{Model} & \textbf{\begin{tabular}[c]{@{}c@{}}Accuracy\\ (\%)\end{tabular}} & \textbf{Precision} & \textbf{Recall} & \textbf{F1-Score} & \textbf{\begin{tabular}[c]{@{}c@{}}Inference \\ Time\end{tabular}} \\ \hline
\multirow{5}{*}{Paddy Leaf} & VGG16 & 96.96 & 0.97 & 0.97 & 0.97 & 0.13 \\
 & ResNet50 & \textbf{99.57} & \textbf{1.00} & \textbf{1.00} & \textbf{1.00} & \textbf{0.11} \\
 & DenseNet169 & 98.87 & 0.99 & 0.99 & 0.99 & 0.13 \\
 & MobileNetV2 & 99.68 & 1.00 & 1.00 & 1.00 & 0.12 \\
 & EfficientNet-B0 & 98.98 & 0.99 & 0.99 & 0.99 & 0.10 \\ \hline
\multirow{5}{*}{PlantVillage} & VGG16 & 95.94 & 0.96 & 0.92 & 0.93 & 0.083 \\
 & ResNet50 & \textbf{99.33} & \textbf{0.99} & \textbf{0.99} & \textbf{0.99} & \textbf{0.090}\\
 & DenseNet169 & 95.16 & 0.94 & 0.90 & 0.90 & 0.086 \\
 & MobileNetV2 & 98.60 & 0.99 & 0.98 & 0.98 & 0.085 \\
 & EfficientNet-B0 & 99.14 & 0.99 & 0.98 & 0.98 & 0.086 \\ \hline
\end{tabular}
\label{tab: t3}
\end{table}
The first step of the compression pipeline applies Knowledge Distillation, significantly reducing model parameters and memory requirements. Table \ref{tab: stud} shows the summary of the light-weight student models.
\begin{table}[]
\centering
\caption{Summary of light-weight student models}

\begin{tabular}{cccc}
\hline
\textbf{Student Models} & \textbf{\begin{tabular}[c]{@{}c@{}}No. of Weighted \\ Parameters\end{tabular}} & \textbf{\begin{tabular}[c]{@{}c@{}}Memory Usage\\ (MB)\end{tabular}} & \textbf{\begin{tabular}[c]{@{}c@{}}Estimated Total Size\\ (MB)\end{tabular}} \\ \hline
VGG-16 & 25.78M & 98.35 & 144.49 \\ 
ResNet50 & 4.1M & 15.66 & 125.25 \\ 
DenseNet169 & 0.285M & 1.09 & 45.46 \\ 
MobileNetV2 & 0.2M & 0.77 & 93.22 \\ 
EfficientNet-B0 & 2.28M & 8.72 & 65.57 \\ \hline
\end{tabular}
\label{tab: stud}
\end{table}
In Table \ref{tab: t3} ResNet50, with 4.10M parameters, maintains high accuracy across datasets, achieving 99.57\% and 99.33\% along with near-perfect precision, recall, F1, and faster inference times. MobileNet, with only 0.20M parameters, achieves 99.68\% and 98.60\% accuracy. EfficientNet also performs well, reaching 98.98\% and 99.14\% accuracy with 2.28M parameters. Though VGG shows comparatively lower accuracy, it surpasses the original VGG16 baseline while reducing parameters to 25.78 M. DenseNet achieves optimal results despite a significant parameter drop to 0.285M, with 98.87\% accuracy on Paddy-Leaf and 95.16\% on PlantVillage. All models show improved inference times, demonstrating the effectiveness of Knowledge Distillation in producing lightweight, accurate models.
\begin{figure}[htbp]
    \centering
    \begin{subfigure}{0.48\linewidth}
        \centering
        \includegraphics[width=\linewidth]{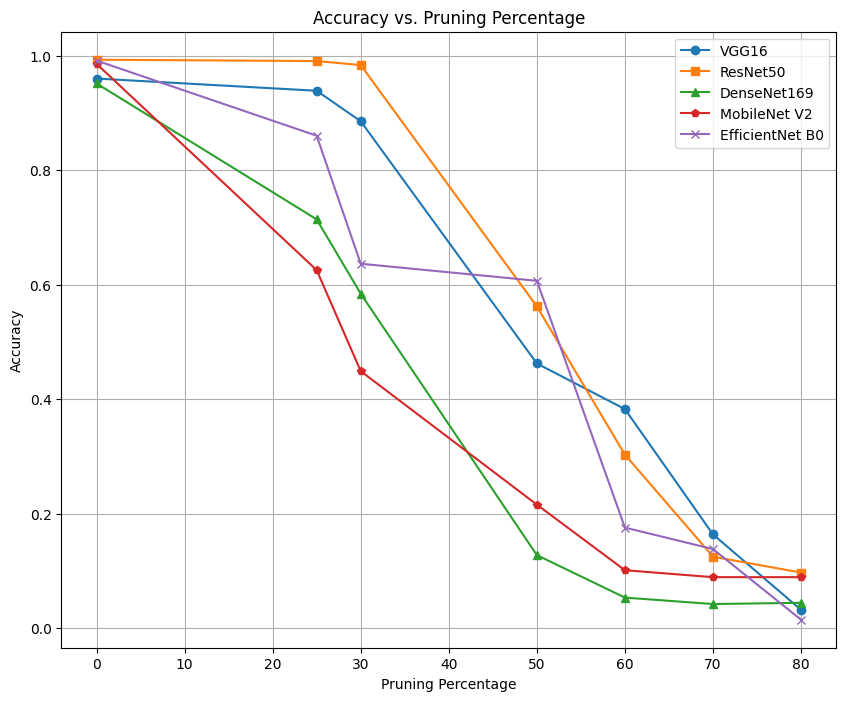}
        \caption{PlantVillage dataset}
    \end{subfigure}
    \hspace{0.5em}
    \begin{subfigure}{0.48\linewidth}
        \centering
        \includegraphics[width=\linewidth]{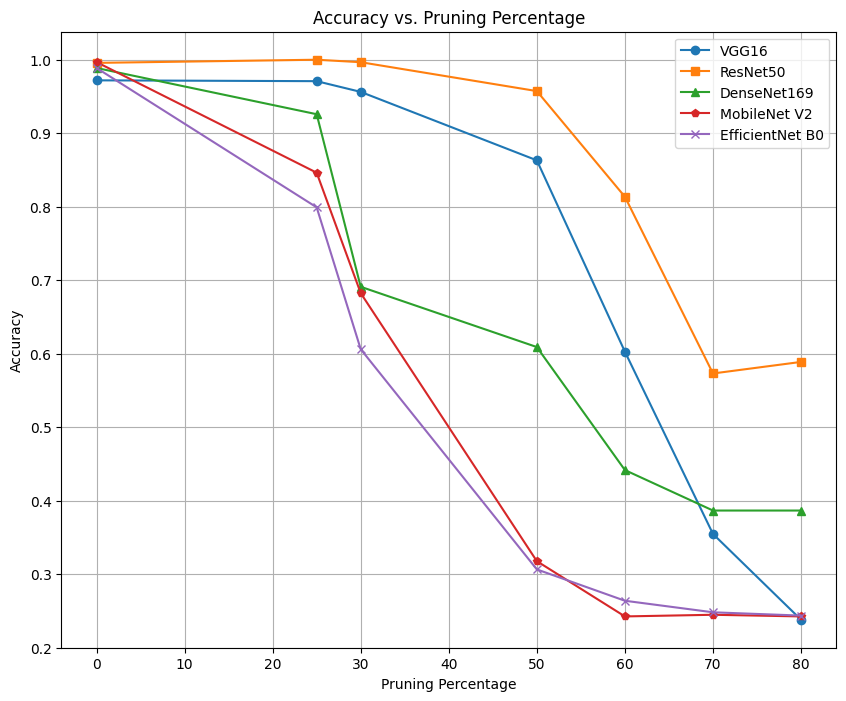}
        \caption{Paddy Leaf dataset}
    \end{subfigure}
    \caption{Accuracy vs pruning percentage curves for the PlantVillage and Paddy Leaf datasets}
    \label{c2}
\end{figure}
\begin{table}[htbp]
\centering
\caption{Results of the models after the second step of two-step compression}
\resizebox{\linewidth}{!}{
\begin{tabular}{cccccccc}
\hline
\textbf{Dataset} & \textbf{Model} & \textbf{\begin{tabular}[c]{@{}c@{}}Optimal Pruning \\ Rate (\%)\end{tabular}} & \textbf{\begin{tabular}[c]{@{}c@{}}No. of Weighted \\ Parameters\end{tabular}} & \textbf{\begin{tabular}[c]{@{}c@{}}Accuracy\\ (\%)\end{tabular}} & \textbf{Recall} & \textbf{F1-Score} & \textbf{\begin{tabular}[c]{@{}c@{}}Inference \\ Time\end{tabular}} \\ \hline
\multirow{3}{*}{Paddy Leaf} & VGG-16 & 30 & 18M & 95.73 & 0.96 & 0.96 & 0.11 \\  
 & ResNet50 & \textbf{30} & \textbf{2.85M} & \textbf{99.55} & \textbf{1.00} & \textbf{1.00} & \textbf{0.11} \\ 
 & DenseNet169 & 10 & 0.257M & 99.21 & 0.99 & 0.99 & 0.11 \\ \hline
\multirow{4}{*}{PlantVillage} & VGG-16 & 15 & 22M & 95.40 & 0.92 & 0.93 & 0.086 \\  
 & ResNet50 & \textbf{25} & \textbf{3.1M} & \textbf{98.99} & \textbf{0.99} & \textbf{0.99} & \textbf{0.089} \\  
 & DenseNet169 & 10 & 0.257M & 95.04 & 0.87 & 0.87 & 0.085 \\ 
 & EfficientNet-B0 & 10 & 2.06M & 98.91 & 0.98 & 0.98 & \textbf{0.085} \\ \hline
\end{tabular}}
\label{tab: t4}
\end{table}
Following step 1, adaptive pruning was applied to further compress the student models. Table \ref{tab: t4} depicts the results of the second step of compression. The accuracy vs. pruning rate curve (Figure \ref{c2}) shows that for the Paddy Leaf dataset, MobileNet and EfficientNet experience significant accuracy drops with increased pruning, suggesting limited compressibility. Similarly, MobileNet is sensitive to pruning on the PlantVillage dataset. ResNet demonstrates strong resilience, maintaining 99.55\% accuracy on the Paddy Leaf dataset after 30\% pruning, reducing parameters to 2.85 M. For PlantVillage, ResNet allows 25\% pruning with minimal performance loss, lowering parameters to 3.1 M. VGG handles 30\% and 15\% pruning for the two datasets, reducing to 18M and 22M parameters, respectively. DenseNet maintains high accuracy on the Paddy Leaf dataset with 0.257M parameters after 10\% pruning. EfficientNet performs well on the PlantVillage dataset, retaining 98.91\% accuracy with 2.06M parameters and consistent recall, F1, and inference time.
\subsection{Ablation Study and Performance Analysis of Hybrid Model}
\begin{table}[h]
\centering
\caption{Results of proposed model}
\resizebox{\linewidth}{!}{\begin{tabular}{cccccccc}
\hline
\textbf{Dataset} & \textbf{\begin{tabular}[c]{@{}c@{}}Compression \\ Step\end{tabular}} & \textbf{Model} & \textbf{\begin{tabular}[c]{@{}c@{}}Accuracy\\ (\%)\end{tabular}} & \textbf{Recall} & \textbf{F1-Score} & \textbf{\begin{tabular}[c]{@{}c@{}}Inference Time\\ (sec)\end{tabular}} & \textbf{\begin{tabular}[c]{@{}c@{}}Weight \\ Parameters\end{tabular}} \\ \hline
\multirow{6}{*}{PlantVillage} & \multirow{3}{*}{\begin{tabular}[c]{@{}c@{}}Knowledge\\ Distillation\end{tabular}} & Hybrid (1 Layer Involution) & 95.91 & 0.87 & 0.87 & 0.092 & 0.29M \\
 &  & Hybrid (2 Layers Involution) & 96.76 & 0.90 & 0.90 & 0.088 & 0.30M \\
 &  & Hybrid (3 Layers Involution) & \textbf{97.1} & \textbf{0.91} & \textbf{0.91} & \textbf{0.088} & \textbf{0.32M} \\ \cline{2-8} 
 & \multirow{3}{*}{\begin{tabular}[c]{@{}c@{}}Weight Pruning\\ (10\%)\end{tabular}} & Hybrid (1 Layer Involution) & 95.61 & 0.87 & 0.87 & 0.089 & 0.26M \\
 &  & Hybrid (2 Layers Involution) & 96.71 & 0.90 & 0.90 & 0.093 & 0.28M \\
 &  & Hybrid (3 Layers Involution) & \textbf{96.99} & \textbf{0.91} & \textbf{0.91} & \textbf{0.089} & \textbf{0.29M} \\ \hline
\multirow{6}{*}{Paddy Leaf} & \multirow{3}{*}{\begin{tabular}[c]{@{}c@{}}Knowledge\\ Distillation\end{tabular}} & Hybrid (1 Layer Involution) & 98.71 & 0.99 & 0.99 & 0.15 & 0.29M \\
 &  & Hybrid (2 Layers Involution) & 98.65 & 0.99 & 0.98 & 0.11 & 0.30M \\
 &  & Hybrid (3 Layers Involution) & \textbf{98.87} & \textbf{0.99} & \textbf{0.99} & \textbf{0.11} & \textbf{0.32M} \\ \cline{2-8} 
 & \multirow{3}{*}{\begin{tabular}[c]{@{}c@{}}Weight Pruning\\ (10\%)\end{tabular}} & Hybrid (1 Layer Involution) & 98.65 & 0.99 & 0.99 & 0.14 & 0.26M \\
 &  & Hybrid (2 Layers Involution) & 98.54 & 0.99 & 0.98 & 0.12 & 0.28M \\
 &  & Hybrid (3 Layers Involution) & \textbf{98.63} & \textbf{0.99} & \textbf{0.99} & \textbf{0.11} & \textbf{0.29M} \\ \hline
\end{tabular}}
\label{tab: t5}
\end{table}
\begin{figure}[]
    \centering
    \begin{subfigure}{0.49\linewidth}
        \centering
        \includegraphics[width=\linewidth]{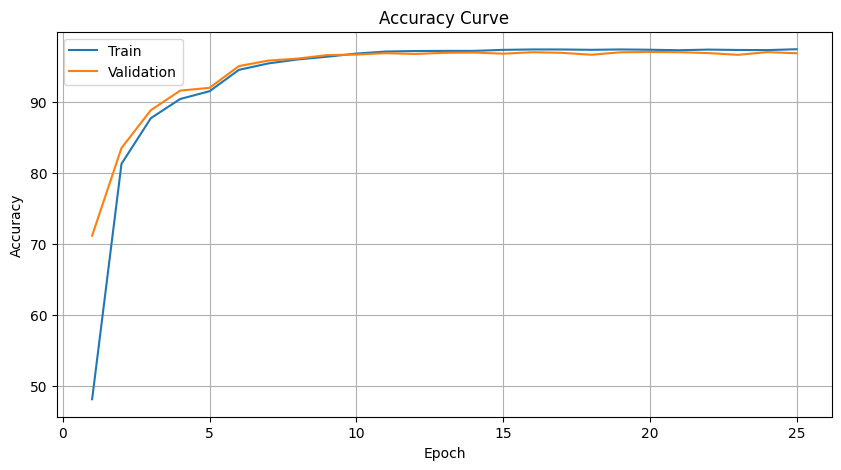}
        \caption{Accuracy curve}
    \end{subfigure}
    \hspace{0.5em}
    \begin{subfigure}{0.48\linewidth}
        \centering
        \includegraphics[width=\linewidth]{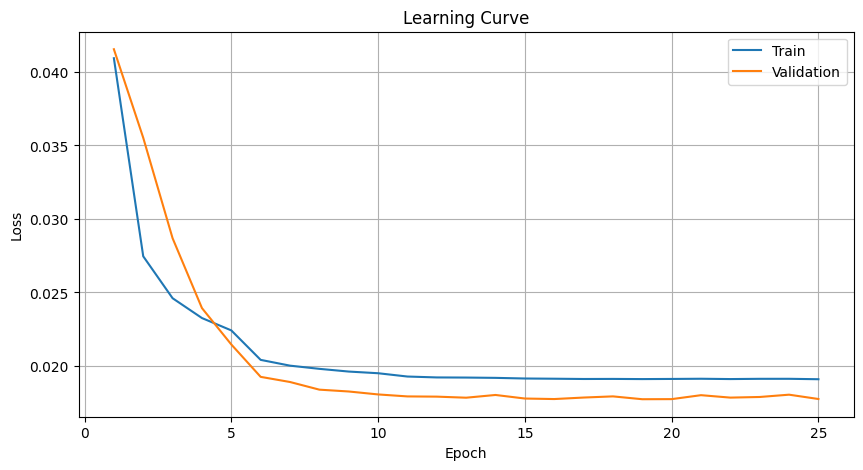}
        \caption{Learning curve}
    \end{subfigure}
    \caption{Accuracy and learning curves of the hybrid model for the PlantVillage dataset}
    \label{c3}
\end{figure}
\begin{figure}[ht]
    \centering
    \begin{subfigure}{0.48\linewidth}
        
    \centering
    \includegraphics[width=\linewidth]{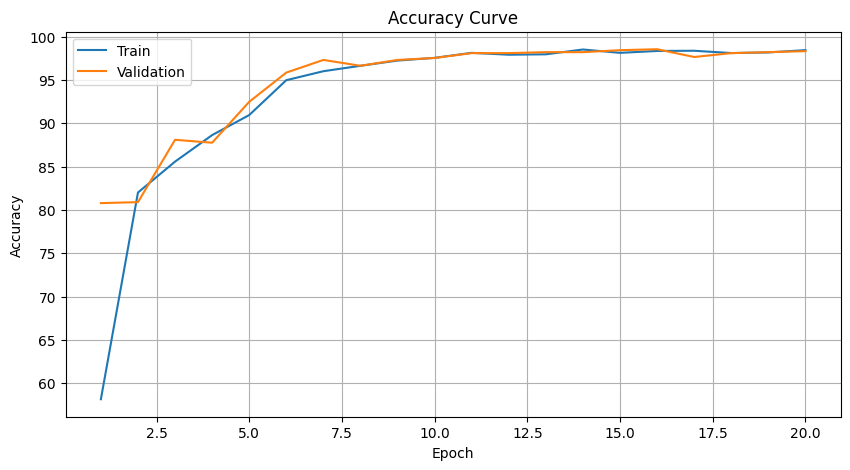}
    \caption{Accuracy curve}
    \end{subfigure}
    \hspace{0.5em}
    \begin{subfigure}{0.49\linewidth}
        
    \centering
    \includegraphics[width=\linewidth]{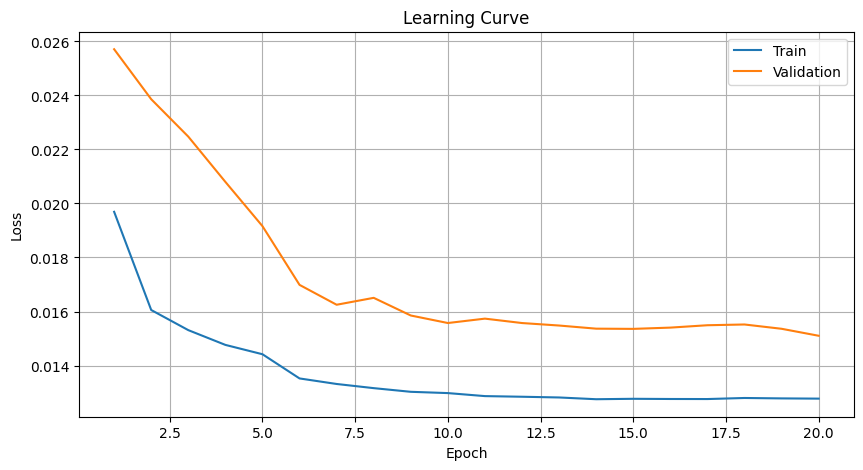}
    \caption{Learning curve}
    \end{subfigure}
    \caption{Accuracy curve and learning curves of the hybrid model for Paddy Leaf dataset}
    \label{c4}
\end{figure}
This section presents the performance of the hybrid student model, which integrates involutional (Inv) layers after DenseBlocks, evaluated across both steps of the compression pipeline. As we can see in Table \ref{tab: t5}, for the Paddy Leaf dataset, during the knowledge distillation step, the hybrid model with one Inv layer achieves 98.71\% accuracy, high recall, an F1 score, and a parameter count of 0.29 million, with an inference time of 0.15 seconds. Adding a second Inv layer slightly lowers the accuracy to 98.65\% but improves inference time to 0.11 seconds. With three Inv layers, the model reaches 98.87\% accuracy while maintaining 0.11-second inference and 0.32 million parameters. After applying 10\% pruning in step 2, all configurations maintain similar performance with reduced parameter counts. 

\begin{figure}[h]
    \centering
    \begin{subfigure}{\linewidth}
        \centering
        \includegraphics[width=\linewidth]{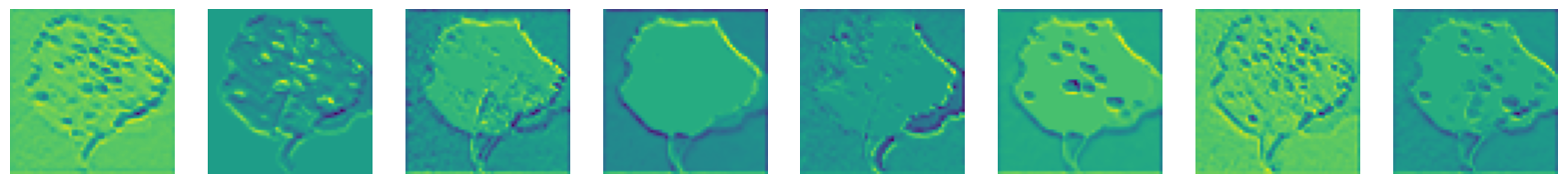} 
        \caption{Layer 1}
    \end{subfigure}
    
    \vspace{2mm} 

    \begin{subfigure}{\linewidth}
        \centering
        \includegraphics[width=\linewidth]{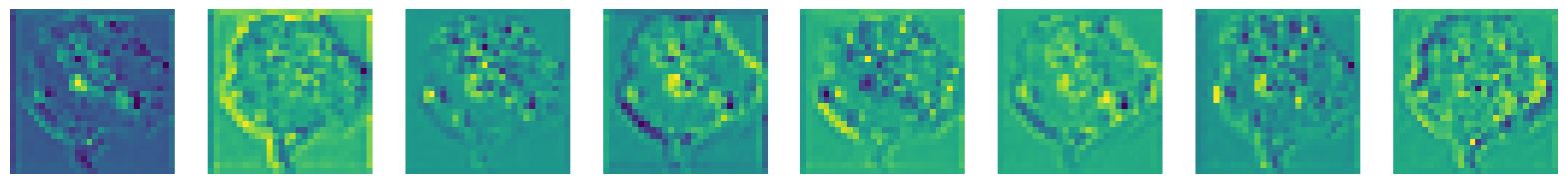} 
        \caption{Layer 2}
    \end{subfigure}
    
    \vspace{2mm} 

    \begin{subfigure}{\linewidth}
        \centering
        \includegraphics[width=\linewidth]{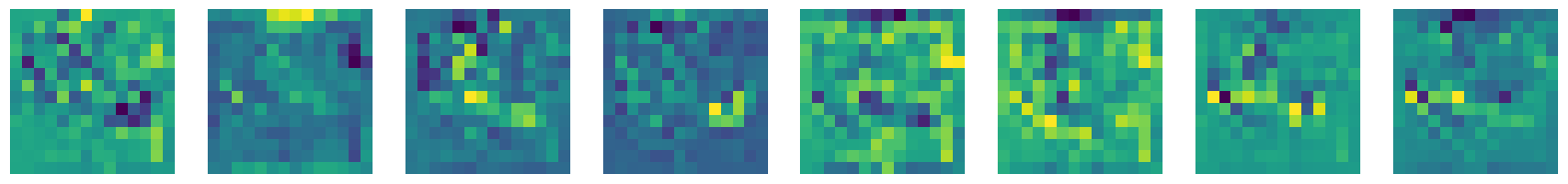} 
        \caption{Layer 3}
    \end{subfigure}

    \caption{ Involution kernel visualization of the hybrid Model for the PlantVillage dataset}
    \label{i2}
\end{figure}

On the PlantVillage dataset, the three-layer Inv model delivers optimal results, achieving 97.10\% accuracy, 0.99 recall, F1 score, and a fast inference time of 0.088 seconds. Accuracy and learning curves for both datasets are shown in Figures \ref{c3} and \ref{c4}, while Figure \ref{i2} visualizes the involutional kernels. These results demonstrate the hybrid model’s ability to balance accuracy, efficiency, and parameter reduction across both datasets and pipeline stages.

\begin{figure}[b]
    \centering
    \includegraphics[width=0.3\linewidth]{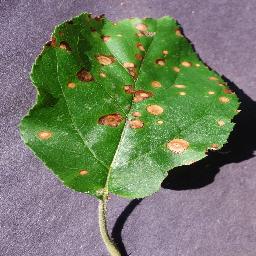}
    \caption{Input image from PlantVillage}
    \label{fig: inp}
\end{figure}
\subsection{Robustness Test}

\begin{table}[ht]
\caption{Result analysis after Noise Attack on both the datasets}
\resizebox{\linewidth}{!}{
\begin{tabular}{|c|c|cccccc|}
\hline
\multirow{2}{*}{\textbf{Dataset}}                                                     & \multirow{2}{*}{\textbf{Parameters}}                                     & \multicolumn{6}{c|}{\textbf{Models}}                                                                                                                                                                                                                         \\ \cline{3-8} 
                                                                                      &                                                                          & \multicolumn{1}{c|}{VGG-16}  & \multicolumn{1}{c|}{ResNet50} & \multicolumn{1}{c|}{DenseNet169} & \multicolumn{1}{c|}{MobileNet\_V2} & \multicolumn{1}{c|}{EfficientNet\_B0} & \textbf{\begin{tabular}[c]{@{}c@{}}Hybrid DenseNet\\ (Proposed)\end{tabular}} \\ \hline
\multirow{2}{*}{\begin{tabular}[c]{@{}c@{}}PlantVillage\\ \&\\ RiceLeaf\end{tabular}} & \begin{tabular}[c]{@{}c@{}}MACS\\ (\num{e9})\end{tabular}  & \multicolumn{1}{c|}{0.50980}  & \multicolumn{1}{c|}{0.803212}  & \multicolumn{1}{c|}{0.335334}     & \multicolumn{1}{c|}{0.282611}       & \multicolumn{1}{c|}{1.453}            & \textbf{0.372898}                                                              \\ \cline{2-8} 
                                                                                      & \begin{tabular}[c]{@{}c@{}}FLOPS\\ (\num{e9})\end{tabular} & \multicolumn{1}{c|}{1.019}    & \multicolumn{1}{c|}{1.6064}   & \multicolumn{1}{c|}{0.670667}     & \multicolumn{1}{c|}{0.565223}       & \multicolumn{1}{c|}{2.906}             & \textbf{0.745796}                                                              \\ \hline
\multirow{5}{*}{PlantVillage}                                                         & \begin{tabular}[c]{@{}c@{}}Accuracy\\ Step 1\end{tabular}                & \multicolumn{1}{c|}{84.44\%} & \multicolumn{1}{c|}{90.20\%}  & \multicolumn{1}{c|}{70.25\%}     & \multicolumn{1}{c|}{75.14\%}       & \multicolumn{1}{c|}{75.69\%}          & \textbf{89.01\%}                                                              \\ \cline{2-8} 
                                                                                      & \begin{tabular}[c]{@{}c@{}}Accuracy\\ Step 2\end{tabular}                & \multicolumn{1}{c|}{84.15\%} & \multicolumn{1}{c|}{84.22\%}  & \multicolumn{1}{c|}{70.10\%}     & \multicolumn{1}{c|}{N/A}           & \multicolumn{1}{c|}{75.74\%}          & \textbf{87.42\%}                                                              \\ \cline{2-8} 
                                                                                      & Precision                                                                & \multicolumn{1}{c|}{0.87}    & \multicolumn{1}{c|}{0.91}     & \multicolumn{1}{c|}{0.77}        & \multicolumn{1}{c|}{0.88}          & \multicolumn{1}{c|}{0.88}             & \textbf{0.88}                                                                 \\ \cline{2-8} 
                                                                                      & Recall                                                                   & \multicolumn{1}{c|}{0.78}    & \multicolumn{1}{c|}{0.86}     & \multicolumn{1}{c|}{0.53}        & \multicolumn{1}{c|}{0.68}          & \multicolumn{1}{c|}{0.65}             & \textbf{0.81}                                                                 \\ \cline{2-8} 
                                                                                      & F1-score                                                                 & \multicolumn{1}{c|}{0.79}    & \multicolumn{1}{c|}{0.86}     & \multicolumn{1}{c|}{0.55}        & \multicolumn{1}{c|}{0.66}          & \multicolumn{1}{c|}{0.69}             & \textbf{0.82}                                                                 \\ \hline
\multirow{5}{*}{RiceLeaf}                                                             & \begin{tabular}[c]{@{}c@{}}Accuracy\\ Step 1\end{tabular}                & \multicolumn{1}{c|}{92.80\%} & \multicolumn{1}{c|}{97.97\%}  & \multicolumn{1}{c|}{95.73\%}     & \multicolumn{1}{c|}{98.09\%}       & \multicolumn{1}{c|}{94.94\%}          & \textbf{97.86\%}                                                              \\ \cline{2-8} 
                                                                                      & \begin{tabular}[c]{@{}c@{}}Accuracy\\ Step 2\end{tabular}                & \multicolumn{1}{c|}{94.83\%} & \multicolumn{1}{c|}{98.87\%}  & \multicolumn{1}{c|}{94.94\%}     & \multicolumn{1}{c|}{N/A}           & \multicolumn{1}{c|}{N/A}              & \textbf{98.76\%}                                                              \\ \cline{2-8} 
                                                                                      & Precision                                                                & \multicolumn{1}{c|}{0.95}    & \multicolumn{1}{c|}{0.99}     & \multicolumn{1}{c|}{0.95}        & \multicolumn{1}{c|}{0.98}          & \multicolumn{1}{c|}{0.95}             & \textbf{0.99}                                                                 \\ \cline{2-8} 
                                                                                      & Recall                                                                   & \multicolumn{1}{c|}{0.95}    & \multicolumn{1}{c|}{0.99}     & \multicolumn{1}{c|}{0.95}        & \multicolumn{1}{c|}{0.98}          & \multicolumn{1}{c|}{0.95}             & \textbf{0.99}                                                                 \\ \cline{2-8} 
                                                                                      & F1-score                                                                 & \multicolumn{1}{c|}{0.95}    & \multicolumn{1}{c|}{0.99}     & \multicolumn{1}{c|}{0.95}        & \multicolumn{1}{c|}{0.98}          & \multicolumn{1}{c|}{0.95}             & \textbf{0.99}                                                                 \\ \hline
\end{tabular}}

\label{tab: t11}
\end{table}
To evaluate the generalization capability and structural resilience of the models under degraded conditions
input conditions, we conducted a robustness test by introducing Gaussian blur noise after each
model had undergone the two-step compression pipeline. Notably, several models, such as MobileNet\_V2 and EfficientNet\_B0, were unable to support further pruning after Knowledge
Distillation due to destabilization or unacceptable performance degradation. Consequently, precision, recall, and F1 scores are reported after the final compression step, each model could successfully support.
From the table \ref{tab: t11}, across both datasets and noise conditions, the proposed Hybrid\_DenseNet exhibits strong
robustness in classification performance while maintaining computational efficiency, a key
requirement for deployment in resource-constrained environments. Most notably, it achieves
top-tier F1 scores (0.82 on PlantVillage and 0.99 on RiceLeaf) with only \num{372.89e6} MACs,
significantly outperforming deeper models like ResNet50 and VGG16 in computational cost
while closely matching their predictive metrics. ResNet50 and VGG16, although strong
performers in terms of raw accuracy, demonstrate notable efficiency-robustness trade-offs:
ResNet50 dropped ~6\% accuracy after Step 2 on the PlantVillage dataset and requires more
than 2× the MACs of Hybrid\_DenseNet. DenseNet169, despite its theoretical architectural
efficiency, suffered a drastic F1-score reduction (0.55 on PlantVillage), indicating its sensitivity
towards adversarial noise after compression. Further, models like MobileNet\_V2 and
EfficientNet\_B0 were unable to undergo pruning beyond distillation, highlighting limited
compression adaptability. Their relatively high precision but low recall indicates a skewed
classification behavior, favoring only a subset of classes as a result of feature loss under noise..
In contrast, Hybrid\_DenseNet maintains a balanced precision (0.88) and recall (0.81) even post-pruning, indicating preservation of class-wise sensitivity under noise corruption. The architectural infusion of involution layers within the DenseNet169 backbone allows spatially
adaptive filtering, involution layers help the network attend to local context, a key factor when
high-frequency disease cues are blurred. This adaptiveness allows the model to outperform
traditional convolution-based architectures in low-quality input scenarios without inflating
computational cost. Overall, the proposed Hybrid\_DenseNet stands out as the most robust and
efficient model post-compression and noise attack, achieving a balanced trade-off across
accuracy, F1-score, and hardware efficiency, making it well-suited for real-world agricultural
applications where bandwidth, compute, and image quality may all be constrained.
\section{Discussion} \label{dis}
\begin{figure}[htbp]
    \centering
    \begin{subfigure}{0.48\linewidth}  
        \centering
        \includegraphics[width=\linewidth]{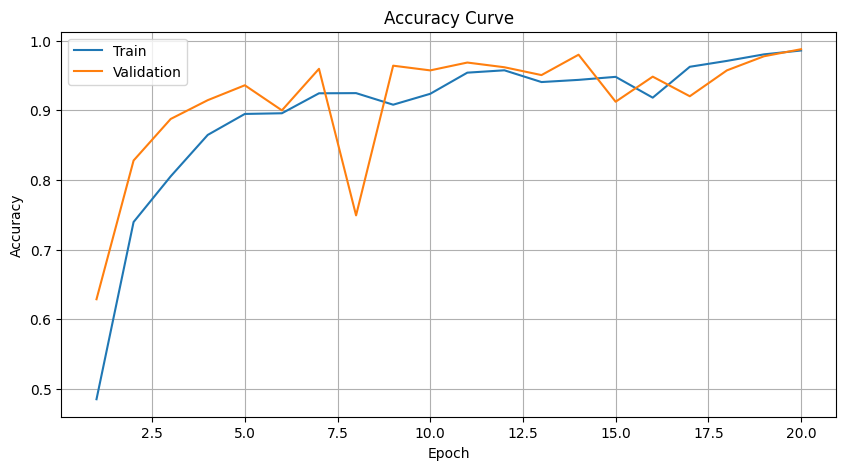}
        \caption{VGG-16 base}
        \label{i4}
    \end{subfigure}
    \hspace{0.5em}  
    \begin{subfigure}{0.48\linewidth}  
        \centering
        \includegraphics[width=\linewidth]{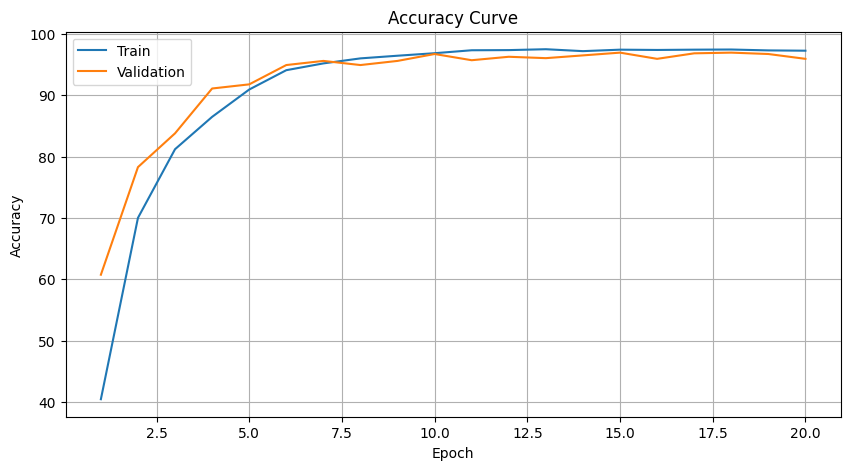}
        \caption{VGG-16 (student)}
        \label{i5}
    \end{subfigure}
    
    \caption{Accuracy curve of VGG-16 baseline and student model on Paddy Leaf dataset}
    \label{i6}
\end{figure}
The ResNet student model retains an accuracy of 99.33\%, which is slightly higher than the performance from the pruning step. But with a parameter count of 4.10 million, it results in a 65\% reduction in terms of parameter counts while maintaining similar or higher accuracy and optimal inference time. By combining these two, after the two-step compression, we can see that the student ResNet50 demonstrates even more compression with a parameter count of 3.1 million only. The evaluation of this compressed model demonstrates an accuracy of 98.99\% with nearly perfect recall and f1 values and an even more optimized inference time of 0.089 seconds. For the other models, we can encounter a similar result, which highlights the effectiveness of the combined compression technique in terms of increasing resource efficiency. 
\begin{table}[ht]
\centering
\caption{A comparative analysis of the base models' performance and compressed models' performance}
\resizebox{\linewidth}{!}{\begin{tabular}{|c|cc|ccc|}
\hline
\multirow{2}{*}{\textbf{Models}} & \multicolumn{2}{c|}{\textbf{Baseline Models}}                                                                      & \multicolumn{3}{c|}{\textbf{Compressed Models}}                                                                                                                     \\ \cline{2-6} 
                                 & \multicolumn{1}{c|}{\textbf{\begin{tabular}[c]{@{}c@{}}Accuracy\\ (\%)\end{tabular}}} & \textbf{Weight Parameters} & \multicolumn{1}{c|}{\textbf{\begin{tabular}[c]{@{}c@{}}Accuracy\\ (\%)\end{tabular}}} & \multicolumn{1}{c|}{\textbf{Weight Parameters}} & \textbf{Compression Step} \\ \hline
VGG-16                           & \multicolumn{1}{c|}{88.14}                                                            & 134M                       & \multicolumn{1}{c|}{95.40}                                                            & \multicolumn{1}{c|}{22M}                        & 2-Step                    \\ \hline
ResNet50                         & \multicolumn{1}{c|}{99.12}                                                            & 23M                        & \multicolumn{1}{c|}{98.99}                                                            & \multicolumn{1}{c|}{3.1M}                       & 2-Step                    \\ \hline
DenseNet169                      & \multicolumn{1}{c|}{99.55}                                                            & 14M                        & \multicolumn{1}{c|}{95.04}                                                            & \multicolumn{1}{c|}{0.257M}                     & 2-Step                    \\ \hline
MobileNetV2                      & \multicolumn{1}{c|}{98.35}                                                            & 2.27M                      & \multicolumn{1}{c|}{98.60}                                                            & \multicolumn{1}{c|}{0.20M}                      & 1-Step                    \\ \hline
EfficientNet-B0                  & \multicolumn{1}{c|}{99.45}                                                            & 4.05M                      & \multicolumn{1}{c|}{98.91}                                                            & \multicolumn{1}{c|}{2.06M}                      & 2-Step                    \\ \hline
Hybrid DenseNet (Proposed)       & \multicolumn{1}{c|}{99.55}                                                            & 14M                        & \multicolumn{1}{c|}{96.99}                                                            & \multicolumn{1}{c|}{0.29M}                      & 2-Step                    \\ \hline
\end{tabular}}
\label{tab: t9}
\end{table}
The overall analysis of this research reveals several important insights into the effectiveness of the involution-infused proposed hybrid student model and the innovative two-step compression pipeline for plant leaf disease classification. The two-step compression process utilizes both Knowledge Distillation and Weight Pruning and demonstrates a strong balance between performance and efficiency. For example, from the experimental analysis, we can see that for the PlantVillage dataset, ResNet50 provides an accuracy of 99.12\% with high precision and an average inference time of 0.13 seconds. 
After applying the post-training pruning on the baseline models, for this particular dataset, ResNet50 retains 98.95\% accuracy with a reduced parameter count of 11.85 million, maintaining high precision and achieving an optimal inference time of 0.091 seconds. If we analyze the effects of only knowledge distillation, we can see after this step.
\begin{figure}[htbp]
    \centering
    \begin{subfigure}{0.48\linewidth}  
        \centering
        \includegraphics[width=\linewidth]{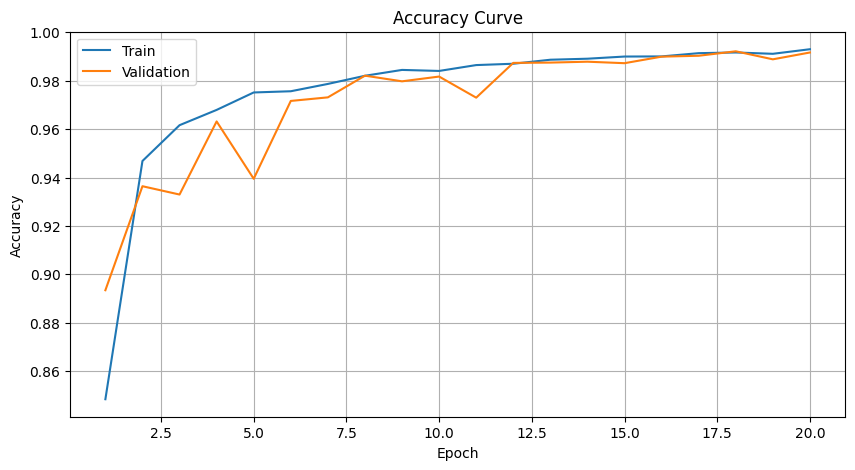}
        
        \caption{ResNet50 base}
        \label{i4}
    \end{subfigure}
    \hspace{0.5em}  
    \begin{subfigure}{0.48\linewidth}  
        \centering
        \includegraphics[width=\linewidth]{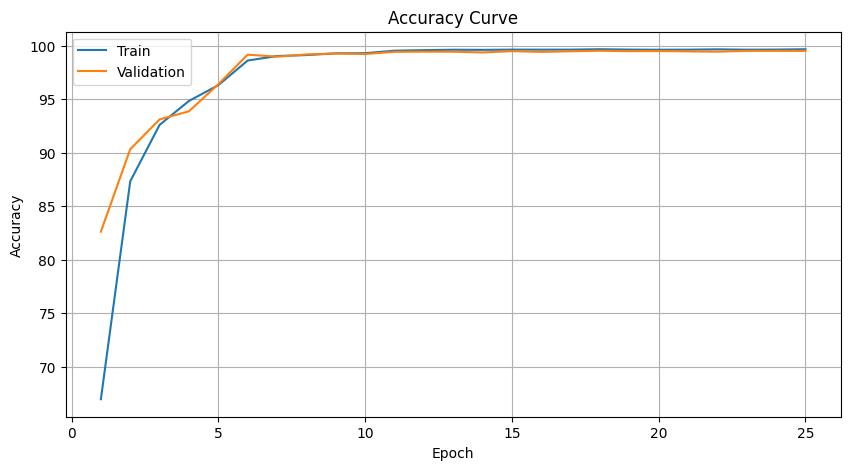}
        \caption{ResNet50 (student)}  
        \label{i5}
    \end{subfigure}
    
    \caption{Accuracy curve of ResNet50 baseline and student model on PlantVillage dataset}
    \label{i9}
\end{figure}

Table \ref{tab: t9} demonstrates a comparative analysis of the base models' performance and compressed models' performance in terms of accuracy and parameter count. While MobileNet achieves slightly better accuracy with fewer parameters, the proposed model demonstrates superior robustness under noise, making it a more reliable choice in adverse conditions. Apart from the compression part, if we look at the first step, we can see that the student model’s training and learning curves demonstrate improved stability compared to the teacher model, as illustrated in Figure \ref{i6} and Figure \ref{i9} for both datasets. This can be attributed to several factors. Firstly, the student model’s simplified architecture, with fewer parameters, allows for efficient optimization and reduces the risk of overfitting, which leads to better regularization. Secondly, during the distillation process, the student models receive the softened outputs, which contain better information than the hard labels, which may help in regularizing the training process. Additionally, the compact structure of the student models allows for efficient parameter utilization, leading to faster convergence and more stability in the learning curve.
\begin{figure}[ht]
    \centering
    \begin{subfigure}{0.48\linewidth} 
        \centering
        \includegraphics[width=\linewidth]{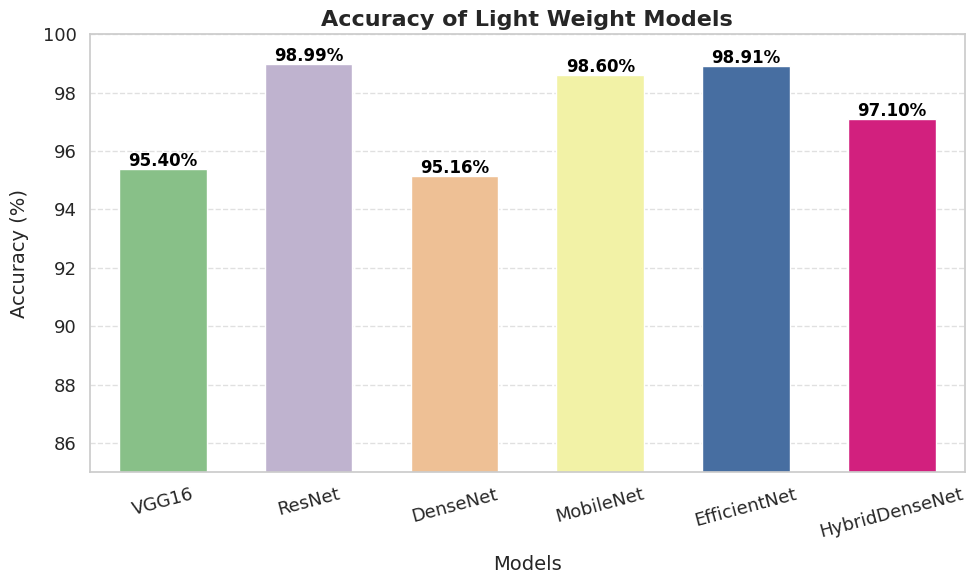}
        \caption{Accuracy of light-weight models}
        \label{i10}
    \end{subfigure}
    \hspace{0.5em} 
    \begin{subfigure}{0.48\linewidth} 
        \centering
        \includegraphics[width=\linewidth]{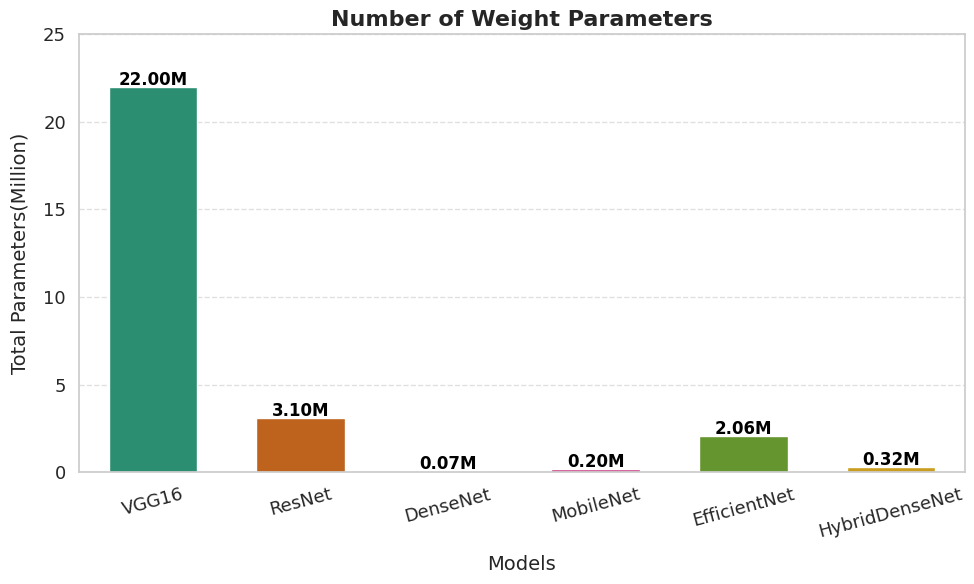}
        \caption{Weighted parameter count of light-weight models}
        \label{i11}
    \end{subfigure}
    
    \caption{Comparative analysis of the performance of the light-weight models}
    \label{i12}
\end{figure}
\begin{figure}
    \centering
    \includegraphics[width=\linewidth]{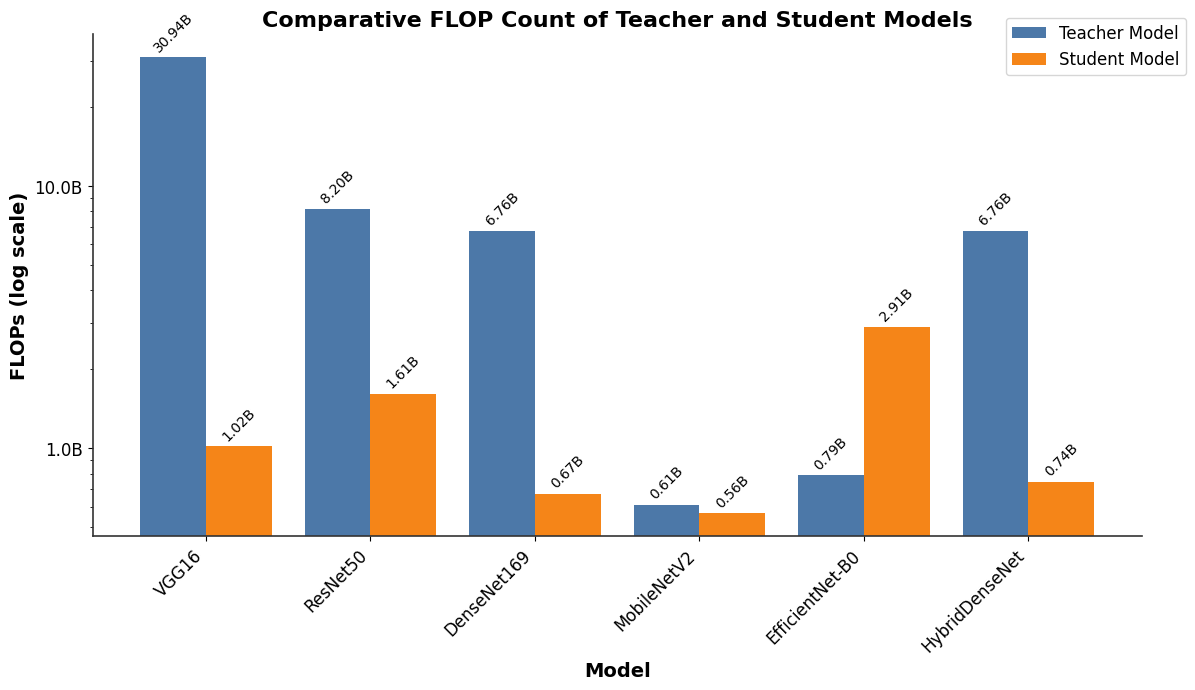}
    \caption{Comparative flop count of Teacher and Student models}
    \label{fig:flop}
\end{figure}
The proposed student hybrid model, which integrates involution layers with DenseBlocks, achieves 
optimal performance with fewer parameters. An ablation study reveals that adding more involution layers improves accuracy while maintaining efficient inference times across various configurations. 
Experimental results indicate that the original DenseNet169 performs consistently well, achieving 99.21%
accuracy on the Paddy Leaf dataset and 93.56\% on the PlantVillage dataset. After two-step compression, 
the student model maintains competitive performance with only 56,000 parameters. Although 
compression slightly reduces performance as parameters are reduced, the model remains optimized 
compared to other architectures, offering room for further improvements. The model’s dense connectivity 
promotes feature reuse, leveraging earlier layers’ features effectively. By incorporating involution layers after DenseBlocks, the network dynamically captures complex spatial relationships, enriching feature extraction. This synergy between adaptive involution layers and dense connectivity results in a powerful model that balances spatial understanding and feature reuse while minimizing parameters. 
A focused analysis of the Floating Point Operations (FLOPs), as shown in Figure \ref{fig:flop} and detailed in Table \ref{tab1} 
and Table \ref{tab: t11}, highlights the computational efficiency achieved through the proposed compression strategy. 

The original VGG16 model requires approximately 30.94 billion FLOPs, the highest among all teacher 
models. After two-step compression, its student variant operates with just 1.02 billion FLOPs, yielding a 
~96.7\% reduction. Similarly, ResNet50 and DenseNet169 are reduced from 8.20B and 6.76B FLOPs to 
1.61B and 0.67B, respectively. The proposed HybridDenseNet achieves a competitive FLOP count of 
0.745B, reflecting an approximate 89\% reduction compared to its teacher model, DenseNet169, while 
retaining high classification accuracy. MobileNetV2 also sees a modest drop from 0.61B to 0.565 B. An 
exception occurs with EfficientNet-B0, where the student model shows an increase in FLOPs (from 0.79B 
to 2.91B) despite halving the parameter count. This is due to the replacement of EfficientNet’s optimized 
mobile blocks with standard convolutions, increasing dense computation. Overall, these FLOP reductions 
(with the exception of EfficientNet-B0) correlate with lower energy consumption and faster inference, 
reinforcing the real-world viability of the proposed two-step compression framework for 
resource-constrained agricultural deployments.
The findings highlight that the proposed hybrid DenseNet model and dual compression strategy strike a 
balance between performance and efficiency. Figures \ref{i10} and \ref{i11} provide a comparative view of the performance and parameter counts for the lightweight models tested on the PlantVillage dataset. Additionally, a recent study has highlighted the growing energy consumption of deep learning algorithms, which surpasses traditional methods and contributes to increased carbon emissions \cite{r35}. The integration of model compression and architectural optimizations in the proposed framework offers a promising energy-efficient solution for plant leaf disease classification, aligning with the broader goal of sustainable AI. The substantial reduction in computational overhead, particularly in FLOP counts, plays a crucial role in lowering energy consumption during inference, making this strategy a step toward more environmentally sustainable deep learning practices. These findings reinforce the practicality of the dual compression approach, offering a pathway to deploy high-performance models on resource-constrained devices with minimal environmental impact. 
\section{Limitations}
In this research, we faced several challenges. The classification report for the hybrid model demonstrates strong performance across most classes, with high precision, recall, and F1 scores. However, classes with fewer samples perform poorly, scoring below 0.40, indicating the need for improved strategies to address class imbalance. The PlantVillage dataset, with its symmetrical samples, consistent backgrounds, and high-quality images, enabled the compressed CNN and hybrid models to perform optimally. Resource constraints also posed challenges. Frequent GPU runtime disconnections on Google Colab initially limited our ability to work with larger datasets. Early experiments using the T4 GPU were conducted on the Paddy Leaf dataset, while the PlantVillage experiments leveraged the A100 GPU for faster results. However, the A100’s high computational cost (10.59 units/hour) restricted broader experimentation, limiting exploration across more diverse datasets.
\section{Conclusion and Future Work}\label{con}
In summary, this research demonstrates the potential of a model compression strategy that combines Weight Pruning, Knowledge Distillation, and hybridization with Involutional Layers to optimize CNNs for crop disease classification. The framework effectively reduces model size and computational complexity, enabling deployment on resource-limited devices like smartphones and edge systems for real-time monitoring. Weight Pruning eliminates redundant parameters while preserving performance, and Knowledge Distillation transfers insights from a larger teacher network to a compact student model, enhancing accuracy. The hybrid model integrates DenseNet with Involutional Layers, providing the student network with the ability to capture complex spatial features using fewer parameters. The effectiveness of the framework is reflected in the accuracy metrics: after compression, ResNet50 achieved 99.55\% and 98.99\% accuracy on the PlantVillage and PaddyLeaf datasets, respectively, while DenseNet recorded optimized accuracies of 99.21\% and 93.96\%. The hybrid model showed competitive performance, achieving 98.87\% and 97.10\% accuracy across the two datasets after the first compression step. Additionally, it showed superior robustness under adverse conditions. These results demonstrate the feasibility of deploying precise, energy-efficient models in agricultural settings, promoting timely disease detection and supporting sustainable crop management. We aim to implement the proposed model in primary datasets in the future for more robust and efficient solutions for real-time disease detection. The proposed framework bridges the gap between cutting-edge machine learning techniques and practical agricultural solutions, paving the way for broader adoption of AI-powered disease diagnosis in farming.
\label{Conclusion and Future Work}

 


\end{document}